\documentclass[sigconf, authorversion, nonacm]{acmart}

\AtBeginDocument{%
  \providecommand\BibTeX{{%
    \normalfont B\kern-0.5em{\scshape i\kern-0.25em b}\kern-0.8em\TeX}}}

\acmPrice{15.00}
\acmISBN{978-1-4503-XXXX-X/18/06}

\usepackage{hyperref}
\usepackage{algorithm}
\usepackage{bbm}
\usepackage{makecell}
\usepackage{algpseudocode}
\usepackage{algpseudocode}
\usepackage{tabularx}
\usepackage{booktabs}
\usepackage{enumitem}
\usepackage{multirow}
\usepackage{colortbl}
\usepackage{subfigure}
\newcommand*{\pr}{\textnormal{Pr}}
\newcommand*{\bt}{\boldsymbol{\theta}}
\newcommand*{\bw}{\boldsymbol{W}}

\begin{document}

\title{Are Your Reviewers Being Treated Equally? Discovering Subgroup Structures to Improve Fairness in Spam Detection}


\author{Jiaxin Liu}
\affiliation{%
 \institution{Lehigh University}
 \country{USA}
 }
\author{Yuefei Lyu}
\affiliation{%
 \institution{BUPT}
 \country{China}
 }
 \author{Xi Zhang}
\affiliation{%
 \institution{BUPT}
 \country{China}
 }
 \author{Sihong Xie}
\affiliation{%
 \institution{Lehigh University}
 \country{USA}
 }





\begin{abstract}
  User-generated reviews of products are vital assets of online commerce, such as Amazon and Yelp, while fake reviews are prevalent to mislead customers. 
GNN is the state-of-the-art method that detects suspicious reviewers by exploiting the topologies of the graph connecting reviewers, reviews, and target products. 
However, the discrepancy in the detection accuracy over different groups of reviewers can degrade reviewer engagement and customer trust in the review websites. 
Unlike the previous belief that the difference between the groups causes unfairness, we study the subgroup structures within the groups that can also cause discrepancies in treating different groups. 
This paper addresses the challenges of defining, approximating, and utilizing a new subgroup structure for fair spam detection. 
We first identify subgroup structures in the review graph that lead to discrepant accuracy in the groups. 
The complex dependencies over the review graph create difficulties in teasing out subgroups hidden within larger groups. 
We design a model that can be trained to jointly infer the hidden subgroup memberships and exploits the membership for calibrating the detection accuracy across groups. 
Comprehensive comparisons against baselines on three large Yelp review datasets demonstrate that the subgroup membership can be identified and exploited for group fairness.
\end{abstract}




\maketitle
\section{Introduction}
Existing fake reviews severely undermine the trustworthiness of online commerce websites that host the reviews~\cite{luca2016reviews}.
People have studied many detection methods to identify those reviews accurately.
Among all methods, graph-based approaches~\cite{rayana2015collective,  li2019spam} have shown great promise.
However, most of the prior methods~\cite{dou2020enhancing, wu2020graph, wang2019semi, dou2020robust}
focused exclusively on either accuracy or robustness of the fraud detectors, ignoring the detection fairness.
Although several existing works studied fairness issues on graph-based classifiers against sensitive attributes, such as gender, age, and race~\cite{kang2020inform, ijcai2019p456, li2021dyadic, ma2021subgroup, dai2021say, agarwal2021towards}. 
Restricted by the anonymity of the spammers, we have no access to the profile of users' attributes to study any fairness problems towards traditional sensitive attributes.
In fact, graph-based spam detectors suffer from another fairness problem: reviewers will receive unfavorable detection based on the number of historical posts (equal to the degree of the reviewer nodes). 
Existing reviewers and spammers receive slack regulations since their few spams are discreetly hidden behind loads of normal content.
In contrast to that, new reviewers who have very few posts face a higher risk of false detection and strict regulation.
Such discrimination harms user trust and leads to less engagement in online commerce activities.
Formally, this type of fairness issue is caused by the variant graph \textit{topologies}~\cite{burkholder2021certification}.
Growing efforts have been dedicated to enhancing fairness towards topological bias~\cite{burkholder2021certification, li2021dyadic, spinelli2021fairdrop, dai2021say, agarwal2021towards}, yet few works of graph-based spam detection task prompt fairness against topological property.

\begin{figure*}
    \centering
    \vspace{-0.3cm}
    \setlength{\abovecaptionskip}{-1pt}
    \includegraphics[width=0.85\textwidth]{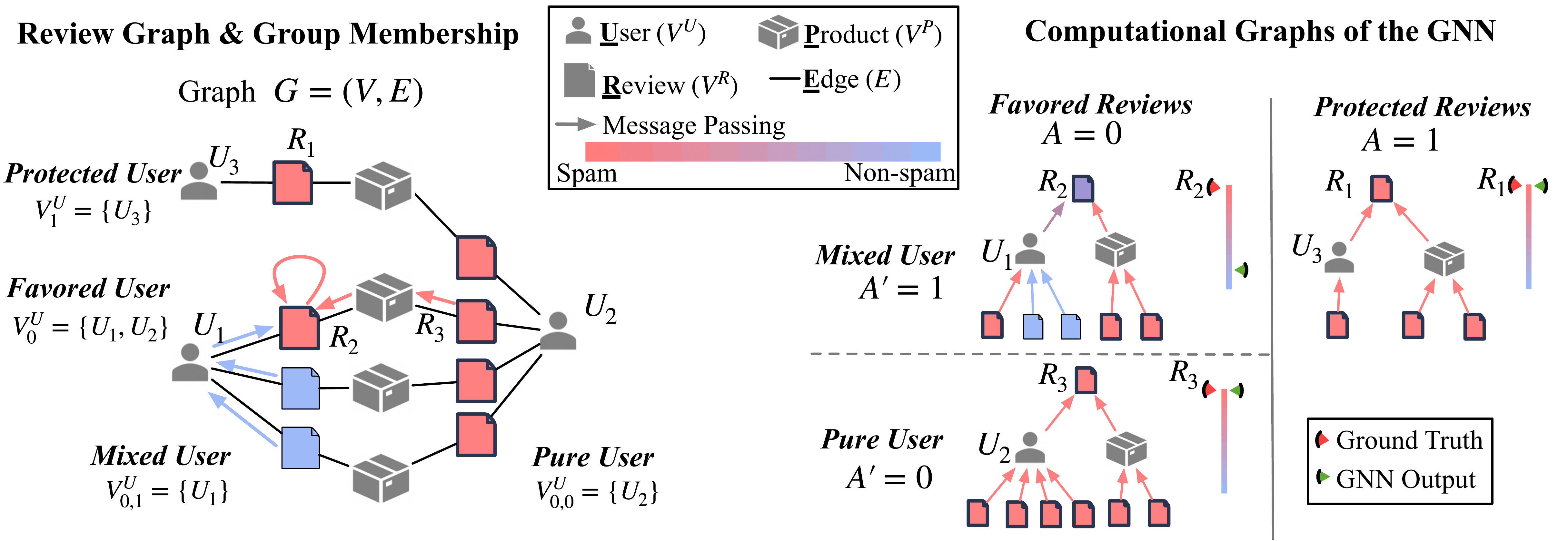}
	\caption{\small\textbf{Problem setting and challenges. Left}: a toy example of the review graph $\mathcal{G}$ and group membership. GNN infers the suspiciousness of reviews (e.g., $R_2$) by passing and aggregating the messages from its neighbors. \textbf{Right}: computational graphs of GNN on spam reviews from different groups and subgroups. GNN unfairly assigns a low suspiciousness to spam $R_2$ posted by mixed user $U_1$, due to the aggregation of messages coming from other non-spams posted by $U_1$.} 
    \label{fig: toy example for the subgroup fairness}
\end{figure*}

Fig.~\ref{fig: toy example for the subgroup fairness} gives an example of our review graph~\cite{luca2016reviews, rayana2015collective} which contains user, review, and product nodes.
Edges represent the events that a user post reviews on some products.
We define users who post fewer reviews than a certain threshold as the ``protected'' users, denoted by the sensitive attribute $A = 1$.
Other users are the "favored" ones, denoted by $A = 0$.
Comparing the computational graphs of spams $R_1$ and $R_2$, many non-spams help dilute the information about the suspiciousness of $R_2$ after messages passing from the bottom up by GNN's aggregation operation in Eq.~(\ref{eq:gnn_agg}).
In this case, existing users with many reviews can reduce suspiciousness and evade detection, which is unfair to new users.

In fact, maintaining fairness between groups of users categorized only by node degree is imprecise.
The detection fairness essentially depends on if two users have the same capability to hide their spamming reviews from their normal reviews. 
Two ``favored'' users can receive different treatment from the detector due to their heterogeneous behaviors, i.e., the various proportions of spams to non-spams across users.
As for ``protected'' users, they post only a few reviews but belong to the same class, reflecting homogeneous user behaviors, whose reviews are treated equally by the detector.
In order to ensure fairness in spam detection, the heterogeneous behavior of ``favored'' users must be accurately described by an additional sensitive attribute $A'$ to indicate if their reviews are from the same class.
In Fig.~\ref{fig: toy example for the subgroup fairness}, user $U_1$, named as the ``mixed'' user, denoted by $A'=1$, posts both spams and non-spams where spams deceive the detector by aggregating messages from non-spams.
$U_2$, named as the ``pure'' user, denoted by $A'=0$, posts only spams and thus receives no messages from non-spams to lower the suspiciousness calculated by the GNN model.. 
In order to get higher accuracy on the favored group, GNN will unfairly target spam reviews posted by pure users like $U_2$ since they are easier to detect. 
Once we let the spams from ``mixed'' users are easy to be caught as well, GNN will have better performance on detecting spams from ``mixed'' users without harming the detection accuracy of spams from ``pure'' user. 
As a result, distinguishing mixed users from pure users is crucial. 
Solving this problem involves three main challenges:

\noindent\textbf{Define the subgroup structure.} 
Most of the previous work studied the fairness problem with well-defined sensitive attributes.
Studies including~\cite{ustun2019fairness, kearns2018preventing} utilize observable sensitive attributes and their combinations to divide the dataset into various groups.
Others~\cite{celis2021fair, awasthi2020equalized, mehrotra2021mitigating} consider the fairness problem with unobserved or noisy sensitive attributes.
In~\cite{chen2019fairness}, researchers built probabilistic models to approximate the value of any well-defined sensitive attribute by using proxy information.
For example, it infers race from the last name and geolocation.
All of these methods are implemented on the I.I.D vector data with well-defined sensitive attributes.
Unlike the above works, we enhance fairness by discovering a novel structural-based sensitive attribute $A'$ and its approximation.
Additionally, sensitive attributes in previous works~\cite{dwork2012fairness, hardt2016equality, zemel2013learning} have been treated as data characteristics, which are not determined by ground truth labels.
We are looking for an undefined sensitive attribute $A'$ that is both specific to graph-based spam detection and related to the ground truth of reviews.
\begin{figure}
    \centering
    \vspace{-0.2CM}
    \setlength{\abovecaptionskip}{-1pt}
    \includegraphics[width=0.4\textwidth]{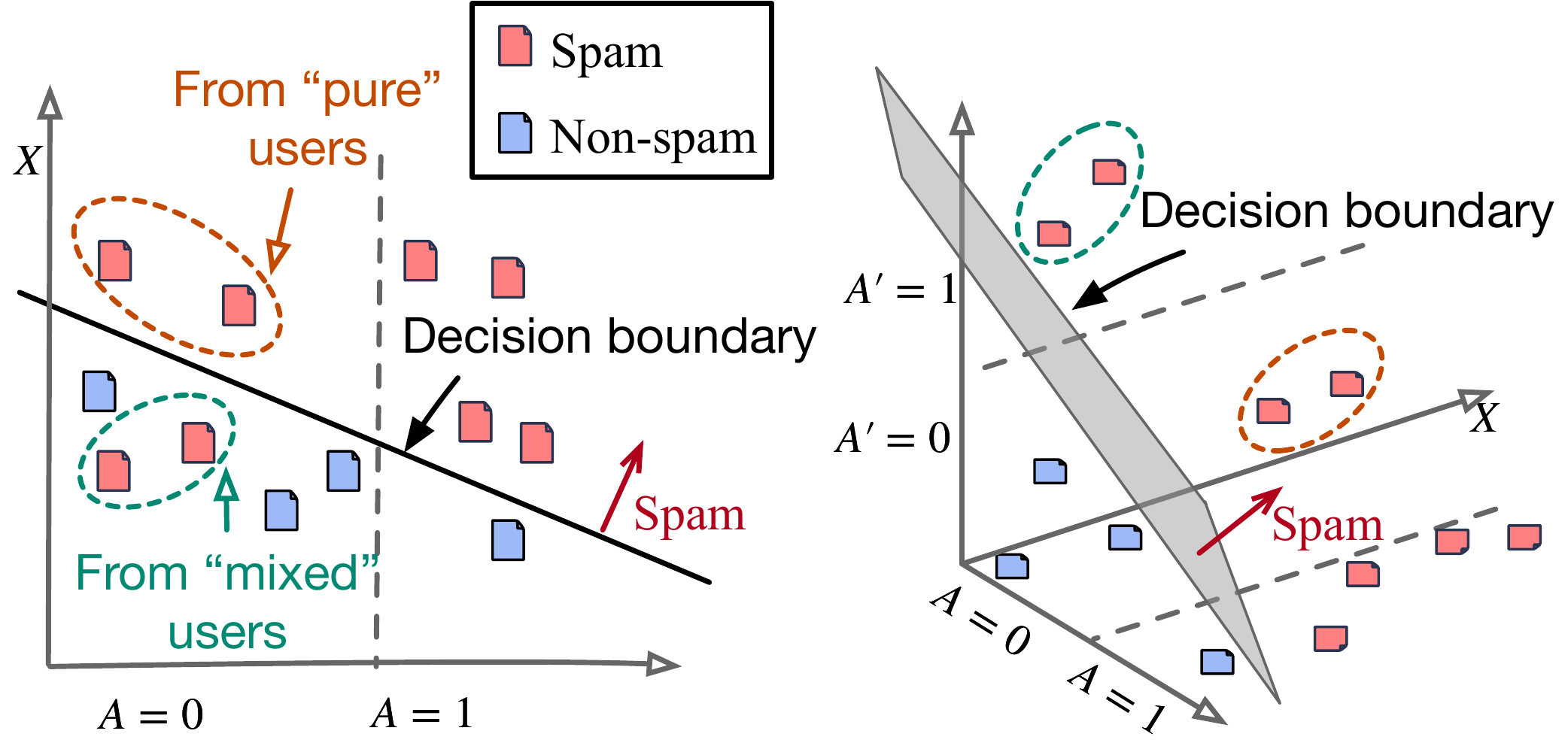}
	\caption{Adding $A'$ improves the accuracy of the detector on spams from the ``favored'' group ($A=0$).}
    \label{fig: adding A'}
	\vspace{-0.3cm}
\end{figure}

\noindent\textbf{Infer unknown subgroup membership.}
We hypothesize that inferring $A'$ helps resolve the unfairness among groups.
Indicating a user's heterogeneous behavior by the subgroup indicator $A'$, either mixed or pure, helps clarify what users from the ``favored'' group genuinely benefit from their non-spams. 
In this case, GNN can use the inferred indicator $A'$ to strike a more equitable detection on spams posted by users from the ``favored'' group and then promote group fairness. (See Fig.~\ref{fig: adding A'}).
Nevertheless, the indicator $A'$ relates to the ground truth labels, which are unobservable for the test data and need to be inferred.
Even for the training data, labeling sufficient data and precisely identifying spams are great challenges that are expensive and time-consuming~\cite{qiu2020adaptive, tan2013unik, tian2020non}. \\
\noindent\textbf{Promote group fairness through subgroup information.} 
Considering fairness towards multiple sensitive attributes, the related works~\cite{kearns2018preventing, ustun2019fairness} formulated the optimization problem with several fairness constraints for each group of combinations. 
These sensitive attributes must be precise and discrete in order to categorize data and ensure that those being discriminated against are fairly treated. 
However, these premises are inapplicable to our inferred subgroup membership indicator $A'$, which is probabilistic rather than deterministic.
Meanwhile, constraints derived from thresholding noisy sensitive attributes can deteriorate optimization algorithms.
Fairness optimization is highly sensitive to group separation, so we avoid setting an uncertain threshold for converting membership from a probability to a concrete group. 

The main contributions of this work are as follows:
 \begin{itemize}[leftmargin=*]
     \item In addition to well-known sensitive attributes, i.e., user node degree~\cite{burkholder2021certification} or node attributes~\cite{bose2019compositional, dai2021say, wu2021learning}, we first define a new structural-based and label-related sensitive attribute $A'$ in the fair spam detection task on graph. 
     \item We propose another GNN $g_{\bt}$ to infer the probability distribution of $A'$ for the test users who have unlabelled reviews (i.e., unobserved user behaviors).
     Due to the insufficient training examples for the ``favored'' group and the ``mixed'' subgroup, we propose two fairness-aware data augmentation methods to synthesize nodes and edges. 
     GNN will generate less biased representations of the minority groups and subgroups by using the re-balanced training datasets. 
     \item Rather than converting the estimated probability distribution of $A'$ into concrete groups by any uncertain threshold, a joint training method is designed to let the detector $f_{\bw}$ directly use the inferred $A'$.
     Our proposed method is evaluated on three real-world Yelp datasets.
     The results demonstrate the unfairness inside the ``favored'' group and show group fairness can be enhanced by introducing the subgroup membership indicator $A'$.
     We show that regardless of what thresholds are used to split the group regarding $A$, our joint method promotes group fairness by accurately inferring the distribution of $A'$ during model training.
 \end{itemize}
\section{Preliminaries}
\subsection{Spam detection based on GNN}
Our spam detection is on the review-graph defined as $\mathcal{G} = (\mathcal{V}, \mathcal{E})$, where $\mathcal{V}=\{v_1, \dots, v_N\}$ denotes the set of nodes and $\mathcal{E} \subseteq \mathcal{V} \times \mathcal{V}$ denotes the set of undirected edges. 
Each node $v_i \in \mathcal{V}$ has a feature vector $\mathbf{x}_i$ with node index $i$.
$\mathcal{G}$ contains three types of nodes: user, review, and product, where each node is from only one of the three types (see the example in Fig.~\ref{fig: toy example for the subgroup fairness} left). 
Let $\mathcal{V}^{U}, \mathcal{V}^{R}$ and $\mathcal{V}^{P}$ denote the sets of user, review, and product nodes, respectively. 
$v_i$ has a set of neighboring nodes denoted by $\mathcal{N}(i) = \{v_j \in \mathcal{V}| e_{i, j}\in \mathcal{E}\}$.
The work focuses on detecting spammy users and reviews, and the task is essentially a node classification problem. 
GNN~\cite{kipf2016semi} is the state-of-the-art method for node classification~\cite{wu2020comprehensive}. 
For our GNN detector $f_{\bw}(\cdot)$, let $\mathbf{h}_i^{(l)}$ be the learned representation of node $v_i$ at layer $l$, where $l=1,\dots L$:
\begin{footnotesize}
\begin{equation}
     \mathbf{h}_i^{(l)} = \text{UPDATE}\left( \text{AGGREGATE}\left(\left\{\mathbf{h}_i^{(l-1)}\right\}\bigcup\left\{{\mathbf{h}_{j}^{(l-1)} \mid j\in \mathcal{N}(i)}\right
    \}\right), \bw^{(l)} \right), \label{eq:gnn_agg}
\end{equation}
\end{footnotesize}
where AGGREGATE takes the mean over $\mathbf{h}_i^{(l)}$ and messages passed from its neighboring nodes.
UPDATE contains an affine mapping with parameters $\boldsymbol{W}^{(l)}$ followed by a nonlinear mapping (ReLU in $l=1,\dots, L-1$ and Sigmoid in $l=L$).
The input vector $\mathbf{x}_i = \mathbf{h}_i^{(0)}$ is the representation at layer 0.
$\hat{y}_i = \mathbf{h}_i^{(L)}\in \mathbb{R} $ denotes the prediction probability of $v_i$ being spam. 
We minimize the cross-entropy loss for the training node set $\mathcal{V}^{\text{Tr}}$:
\begin{small}
\begin{equation}
    \mathcal{L}(\boldsymbol{W}; \mathcal{G}) = \frac{1}{|\mathcal{V}^{\text{Tr}}|} \sum_{v_i \in \mathcal{V}^{\text{Tr}}} \left(- y_i\cdot \log \hat{y}_i - (1 - y_i)\cdot \log(1-\hat{y}_i)\right), \label{eq: cross-entropy loss}
\end{equation}
\end{small} 
where $y_i\in \{0, 1\}$ is the class for $v_i$.
$\bw$ represents a collection of parameters in all the layers $L$.
The main notations are in Table~\ref{tab: notations}.
\subsection{Fairness regularizer}\label{subsec: fairness group}
\noindent\textbf{Group.} 
We split users $\mathcal{V}^U$ into the protected group $\mathcal{V}^U_{1}$ whose degree is
smaller than the $p$-th percentile of all the users' degrees, and the favored group $\mathcal{V}^U_{0}$ for the remaining users (see Fig.~\ref{fig: toy example for the subgroup fairness} left).
The subscript denotes the value of $A$. 
The review nodes $\mathcal{V}^R$ are divided into $\mathcal{V}^R_{1}$ and $\mathcal{V}^R_{0}$ following the group of their associated users.
The fairness of the GNN detector is evaluated by the detection accuracies between two groups of spams $\mathcal{V}^R_0$ and $\mathcal{V}_1^R$.

\noindent\textbf{Fairness regularizer.} Ranking-based metrics like NDCG are appropriate to evaluate the detector's accuracy, as the highly skewed class distribution: most reviews are genuine.
A larger NDCG score indicates that spams are ranked higher than non-spams, and a detector is more precise.
NDCG can also be evaluated on the two groups $\mathcal{V}^R_0$ and $\mathcal{V}^R_1$, separately.
Hence, the group fairness can be evaluated by the NDCG gap between $\mathcal{V}^R_0$ and $\mathcal{V}^R_1$, denoted as $\Delta_{\text{NDCG}}$. 
In fact, NDCG on the ``favored'' group $\mathcal{V}^R_0$ is always lower than that on the $\mathcal{V}^R_1$, since the detector performs preciser on group $\mathcal{V}^R_1$.
For reducing $\Delta_{\text{NDCG}}$ by promoting the NDCG of $\mathcal{V}^R_0$ without hurting that of $\mathcal{V}^R_1$, our detector $f_{\boldsymbol{W}}(\cdot)$ starts with a fairness regularizer $\mathcal{R}_{\text{fair}}$ which takes negative NDCG of $\mathcal{V}^R_0$. 
We adopt a differentiable surrogate NDCG~\cite{burkholder2021certification} for $\mathcal{R}_{\text{fair}}$
\begin{small}
\begin{align}
\setlength{\abovedisplayskip}{4pt}
\setlength{\belowdisplayskip}{4pt}
    \mathcal{R}_{\text{fair}} (\boldsymbol{W}; \mathcal{G}) &=-\frac{1}{Z_0} \sum_{i, j: y_i < y_j \atop v_i, v_j \in \mathcal{V}^{R}_0 \cap \mathcal{V}^{\text{Tr}}} \log \left(1 + \exp \left(\mathbf{h}_j^{(L)} -   \mathbf{h}_i^{(L)}\right)\right), \label{eq: fairness reg} \\[-3pt] 
    \mathcal{L}_{\text{GNN}} (\boldsymbol{W}; \mathcal{G}) &=\mathcal{L}(\boldsymbol{W}; \mathcal{G}) + \lambda \cdot  \mathcal{R}_{\text{fair}} (\boldsymbol{W}; \mathcal{G}), \label{eq: fairGNN loss}
\end{align}
\end{small}
 where $Z_0$ is the total number of pairs of spams and non-spams in the training $\mathcal{V}^R_0$. 
$\mathcal{L}_{\text{GNN}}$ is the objective function for training $f_{\boldsymbol{W}}(\cdot)$ by adding $\mathcal{R}_{\text{fair}}$ to $\mathcal{L}(\boldsymbol{W}; \mathcal{G})$ in Eq.~(\ref{eq: cross-entropy loss}). 
$\lambda>0$ is the importance of the fairness regularizer.
Note that the fairness regularizer $\mathcal{R}_{\text{fair}}$ regularizes all the models referred to in this paper below.
\begin{table}
\small
\vspace{-0.3cm}
\setlength{\abovecaptionskip}{-1pt}
\renewcommand\arraystretch{1.1}
\caption{Notations and definitions.}
\setlength{\tabcolsep}{7pt}
\begin{tabular}{m{1.5cm}<{}m{5.5cm}<{}}  
\toprule
\multicolumn{1}{m{1.5cm}}{\textbf{Notations}} & \multicolumn{1}{m{5.5cm}}{\textbf{Definitions}}\\
\midrule
\multicolumn{2}{l}{\textbf{Graph notations}} \\
\specialrule{0em}{1pt}{1pt}
$\mathcal{G}$ & Review graph \\
$\mathcal{V}, \mathcal{E}$ & Nodes and Edges of graph $\mathcal{G}$ \\
$\mathbf{x}_i, y_i$ & Feature and label of node $v_i$ \\
$\mathcal{N}(i)$ & Set of direct neighbors of $v_i$  \\
$|\mathcal{V}|$ & Cardinality of a set $\mathcal{V}$ \\
$\mathcal{V}^{\text{Tr}}, \mathcal{V}^{\text{Te}}$ & Training nodes and test nodes \\
$\mathcal{V}^U, \mathcal{V}^R, \mathcal{V}^P$ & User, review, and  product nodes \\
\specialrule{0em}{3pt}{3pt}
\multicolumn{2}{l}{\textbf{Group notations}} \\
\specialrule{0em}{1pt}{1pt}
$A, A'$ & Binary sensitive attributes (0/1)\\
$\mathcal{V}^R_a, \mathcal{V}^U_a$ & Review and user nodes from group of $A=a$\\  
$\mathcal{V}^{R}_{a, a'}, \mathcal{V}^{U}_{a, a'}$ & Review and user nodes from group of $A=a$ and $A'=a'$\\
\specialrule{0em}{3pt}{3pt}
\multicolumn{2}{l}{\textbf{Model notations}}\\
\specialrule{0em}{1pt}{1pt}
$f_{\boldsymbol{W}}(\cdot), g_{\boldsymbol{\theta}}(\cdot)$ & GNNs with parameters $\boldsymbol{W}$ and $\boldsymbol{\theta}$ \\ 
$\hat{y}_i, \hat{A}'_i$ &  Output of $f_{\boldsymbol{W}}(\cdot), g_{\boldsymbol{\theta}}(\cdot)$ for $v_i$\\
$\textbf{h}_{i}^{(l)}$ & Representation of $v_i$ on layer $l$\\
$\tilde{\mathbf{x}}_{i j}$ & Synthetic node by mixing-up $v_i$ and $v_j$ \\
$\tilde{y}_{i j}$ & Label for the synthetic node $\tilde{\mathbf{x}}_{i j}$\\
\bottomrule
\end{tabular}
\label{tab: notations}
\vspace{-0.3cm}
\end{table}
\section{Methodology}
\subsection{Subgroup definition and selection}\label{sec 3.1: subgroup definition and inference}
\textbf{Subgroups.}
\begin{figure*}
\vspace{-0.3cm}
    \centering
    \setlength{\abovecaptionskip}{-0.5pt}
    \includegraphics[width=0.85\textwidth]{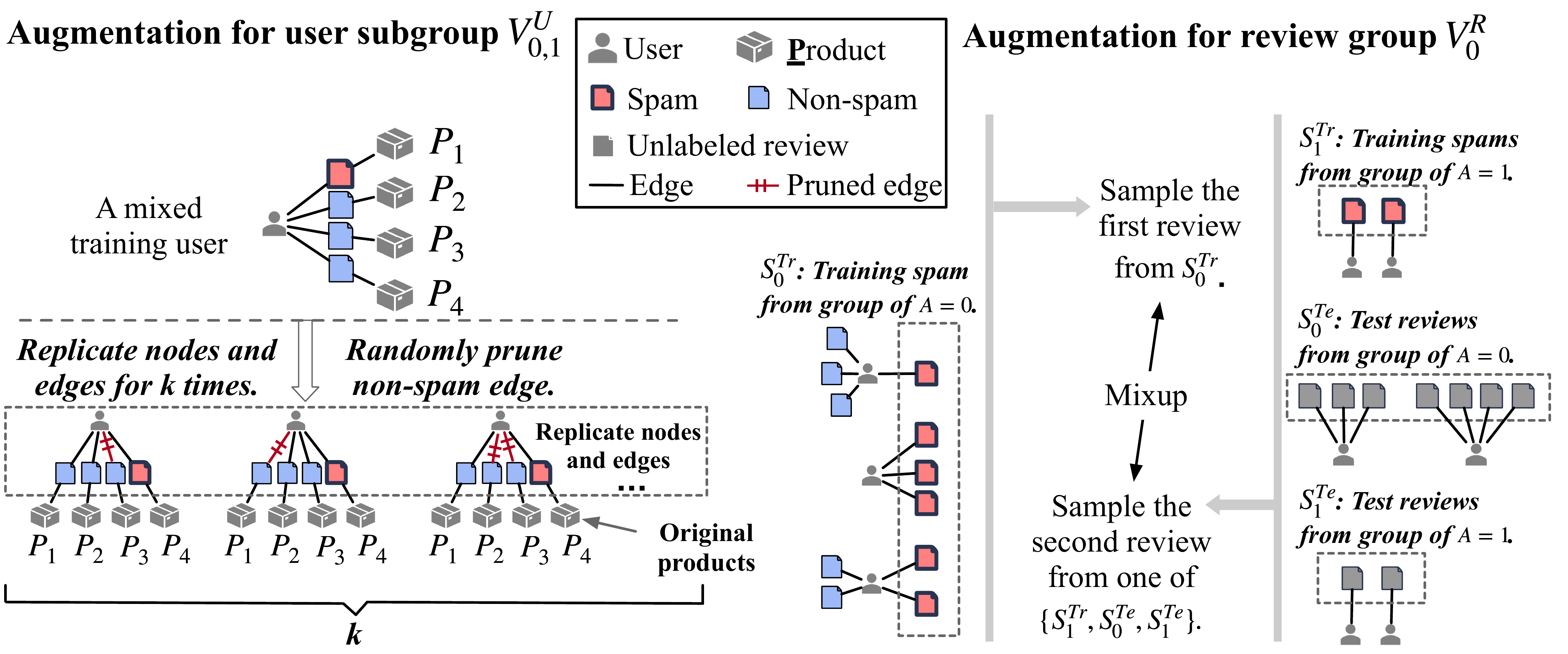}
	\caption{Toy example of our proposed augmentation methods.} 
 \vspace{-0.3cm}
    \label{fig: augmentation}
\end{figure*}
Due to the fact that the suspiciousness of spam posted by user $v_i$ will be decreased as a result of GNN aggregating non-spam posted by user $v_i$.
We define a new sensitive attribute $A'_i$ to identify if user $v_i$ will benefit from its non-spams reviews:
\begin{equation}\label{eq. definition for A'}
    A'_i = \left\{
    \begin{array}{rcl}
        1   & &{\text{if\ } 0 <\sum_{j\in \mathcal{N}(i)} {y_j} < \left| \mathcal{N}(i) \right|} \\
        0   & &{\text{otherwise}} \\
    \end{array}
    \right.
\end{equation}
\noindent where $v_{i}\in \mathcal{V}^U_0$.
$\mathcal{N}(i)$ are reviews posted by $v_i$.
The user with $A'=1$ has both spam ($y=1$) and non-spam ($y=0$) reviews.
$A'=0$ indicates a user posts only reviews that belong to either spams or non-spams but not both.
The heterogeneous behavior of the favored users leads to the split of $\mathcal{V}^U_0$ into two subgroups $\mathcal{V}^U_{0,0}$ and $\mathcal{V}^U_{0, 1}$, where the first and second terms in the subscript represent the value of $A$ and $A'$, respectively.
Since almost all users from $\mathcal{V}^U_1$ post reviews in just one class, $A^\prime$ is no need where $A=1$.

\noindent \textbf{Infer unobservable subgroup membership.} 
We expect the new $A^\prime$ can help the detector to calibrate the prediction over subgroups and promote group fairness.
We can determine the value of $A'$ for users with known classes of reviews.
The majority of review labels are, however, unknown in practice.
Therefore, we use a GNN $g_{\bt}(\cdot)$ to infer $A'$ for users whose reviews are not fully labeled.
In principle, many predictive models can map $v_{i}$ to $A^\prime_i$. 
Still, we choose GNN for its capability of modeling neighborhood data distribution to reflect the heterogeneous properties of users.
Let $\hat{A}'_i$ ($A'_i$, resp.) be the prediction (ground truth, resp.) of $A^\prime$ for user $v_i$.
The loss of predicting $A'_i$ using $g_{\boldsymbol{\theta}}$ becomes:
\begin{small}
\begin{equation}
\setlength{\abovedisplayskip}{4pt}
\setlength{\belowdisplayskip}{4pt}
    \mathcal{L}(\boldsymbol{\theta}; \mathcal{G}) = \frac{1}{Z_1} \sum_{v_i \in \mathcal{V}^{\text{Tr}} \cap \mathcal{V}^U} (- A'_i \cdot \log \hat{A}'_i  - (1 - A'_i)\cdot \log (1 - \hat{A}'_i)), \label{eq: cross-entropy loss for g}
\end{equation}
\end{small}
where $Z_1$ is the total number of training user nodes.
\subsection{Data augmentation for groups and subgroups} \label{sec 3.2 data augmentation}
\subsubsection{Augmentation for user subgroup $\mathcal{V}^U_{0, 1}$.}
\label{sec:minor_user_augmentation}
Since $g_{\boldsymbol{\theta}}$ is under-fitted due to the insufficient training mixed users in our dataset (see Table \ref{tab:dataset}), we synthesize nodes to augment mixed users and their reviews.

First, the synthetic nodes are required to mimic the original distribution of nodes in $\mathcal{V}^U_{0, 1}$.
Oversampling~\cite{chawla2002smote} can straightforwardly add multiple copies of the minority without altering the original node distribution in $\mathcal{V}^U_{0, 1}$.
We \textbf{replicate} the mixed training users ($\mathcal{V}^U_{0, 1}\cap \mathcal{V}^{\text{Tr}}$) along with their reviews for $k$ times, then connect them back to the same products.

Rather than utilizing the exact replications, we slightly perturb the copies.
Unlike the prior augmentation works for image data, such as flipping~\cite{krizhevsky2017imagenet}, cropping~\cite{perez2017effectiveness}, rotation~\cite{cubuk2019autoaugment}, and noise injection~\cite{bishop1995training}, edge pruning can be seen as a unique technology for graph data.
By pruning the review connections of replicated users, the distributions of their neighboring nodes are extended to contain more variations.
Considering the imbalanced classes of reviews from the mixed users: the non-spams dominate all reviews; we randomly \textbf{prune} edges connected to the non-spam reviews and always keep edges connected to the scarce spams (see Fig.~\ref{fig: augmentation} left).
Synthetic mixed users still have multiple reviews (unchanged $A=0$) from two classes (thus preserving $A'=1$).
Therefore, GNN generates diverse but similar representations of synthetic mixed users to the original users.
\subsubsection{Augmentation for minority review group $\mathcal{V}^R_0$.}\label{sec:minor_review_augmentation}
It is challenging to train the detector $f_{\boldsymbol{W}}$ with a limited number of favored reviews (refer to $\% \mathcal{V}^{R}_0$ in Table \ref{tab:dataset}).
Mixup~\cite{zhang2017mixup}, one of the augmentation methods, applies the convex combinations between any two labeled data to interpolate the original sparse distribution.
We augment the minority group $\mathcal{V}^R_0$ following a graph mixup method~\cite{wang2021mixup} that considers both the node features and topology structures. 
The basic mixup for GNN inputs, node embeddings at each layer, and labels for the synthetic data become:
\begin{footnotesize}
\begin{align*}
    \tilde{\mathbf{x}}_{ij} = \alpha \mathbf{x}_i + (1 - \alpha) \mathbf{x}_j, \quad \tilde{\mathbf{h}}_{ij}^{(l)}  = \alpha \tilde{\mathbf{h}}_{i}^{(l)} + (1-\alpha)\tilde{\mathbf{h}}_{j}^{(l)},\quad
    \tilde{y}_{ij} &= \alpha y_i + (1 - \alpha) y_j,
\end{align*}
\end{footnotesize}
where $\alpha \in [0, 1]$ and $\tilde{\mathbf{x}}_{ij}$ is the mixture of node attributes $\mathbf{x}_i$ and $\mathbf{x}_j$ at the input layer.
$\tilde{\mathbf{h}}^{(l)}_{ij}$ is the mixture at the $l$-th layer synthesized from the two hidden representations $\tilde{\mathbf{h}}_{i}^{(l)}$ and $\tilde{\mathbf{h}}_{j}^{(l)}$ ($\tilde{\mathbf{h}}^{(0)}_{ij} = \tilde{\mathbf{x}}_{ij}$).
$\tilde{y}_{ij}$ is the label for $\tilde{\mathbf{x}}_{ij}$. 

Unlike their work, our method carefully delimits nodes for the mixup so that the synthetic data will mitigate the imbalance between the majority and the minority groups or subgroups in our review graph.
Specifically, the first node in our mixing-up method is sampled from the spam reviews in the ``favored'' group $\mathcal{V}_0^R$.
This ensures that the synthetic reviews bear a resemblance to the original favored spams and effectively addresses the issue of class imbalance in the ``favored'' group.
There are three sets for sampling the second nodes where the first set ($S_{1}^{\text{Tr}}$) is our method and the other two sets ($S_{0}^{\text{Te}}$ and $S_{1}^{\text{Te}}$) are variants (see Fig.~\ref{fig: augmentation} right). 
$S_{1}^{\text{Tr}}$ contains spams from the protected training group.
$S_{0}^{\text{Te}}$ contains test reviews from the favored group.
$S_{1}^{\text{Te}}$ contains test reviews from the protected group.
The sets for sampling the first node and the second node are defined as follows:
\begin{small}
\begin{align}
    &\textbf{Sample first node from: } 
    S_{0}^{\text{Tr}} = \left\{v_i \mid v_i \in \mathcal{V}^R_0 \cap \mathcal{V}^{\text{Tr}}, y_i = 1 \right\}. \label{eq: mixup first node}  
    \\
    &\textbf{Sample second node from one of: } S_{1}^{\text{Tr}} = \left\{v_j \mid v_j \in \mathcal{V}^R_1 \cap \mathcal{V}^{\text{Tr}}, y_j = 1 \right\},\notag
    \\
    & S_{0}^{\text{Te}} = \left\{v_j \mid v_j \in \mathcal{V}^R_0 \cap \mathcal{V}^{\text{Te}} \right\}, \quad  S_{1}^{\text{Te}} = \left\{v_j \mid v_j \in \mathcal{V}^R_1 \cap \mathcal{V}^{\text{Te}}\right\}. \label{eq: mixup second node}
\end{align}
\end{small}
Since the labels of reviews sampled from $S_{0}^{\text{Te}}$ and $S_{1}^{\text{Te}}$ are unknown, the synthetic review has the same label as the first node, i.e., $\tilde{y}_{i j} = 1$.
\subsection{Joint model}\label{sec 3.3: joint training}
\subsubsection{Utilizing subgroup membership.} If subgroup membership is known, the most straightforward way is to incorporate an extra fairness regularizer into the objective function that operates on the subgroups defined by the inferred membership. 
However, fairness regularizers require deterministic rather than probabilistic group membership, such as the probabilistic estimation of $A'$ given by $g_{\bt}$.
Hence, we treat the inferred $\hat{A}'$ as an supplementary attribute that informs the detector about the subgroup membership and the uncertainty in the estimation of this membership.
In this case, the expanded feature vector of a user now becomes $\textbf{x}' = [\textbf{x}, \hat{A}']\in \mathbb{R}^{d+1}$ where $\hat{A}'$ is the probability of the user having $A'=1$ given by $g_{\boldsymbol{\theta}}$.
\subsubsection{Optimization for the joint model.}
We propose to optimize two GNNs $g_{\boldsymbol{\theta}}$ and $f_{\boldsymbol{W}}$ jointly so that two GNNs can \textit{co-adopt} to each other. 
As a part of the input to $f_{\boldsymbol{W}}$, $\hat{A}'$ will be treated as a function of $\bt$, i.e., $\hat{A}'(\bt)$, not a constant.
The expanded user feature becomes $\textbf{x}'(\boldsymbol{\theta}) = [\textbf{x}, \hat{A}'(\bt)]$, which involves $\boldsymbol{\theta}$ as parameters.
 In this case, the loss for optimizing the detector $f_{\boldsymbol{W}}$ in Eq. (\ref{eq: fairGNN loss}) becomes $\mathcal{L}_{\text{GNN}}(\boldsymbol{W}, \boldsymbol{\theta}; \mathcal{G})$.
After connecting the computational graph between $\bw$ and $\bt$, $\bt$ receives the gradient from $f_{\bw}$ to enhance the accuracy of inferring $A'$.
Meanwhile, the updated $\bt$ is employed to ameliorate the detector $f_{\bw}$.
During the training, $\bw$ and $\bt$ are updated simultaneously:
\begin{small}
\begin{align}
    \boldsymbol{W} &\gets \boldsymbol{W} - \beta_1 \nabla_{\boldsymbol{W}} \mathcal{L}_{\text{GNN}}(\boldsymbol{W}, \boldsymbol{\theta}; \mathcal{G}) \label{eq: update W} \\
    \boldsymbol{\theta} &\gets \boldsymbol{\theta} -   \underbrace{\beta_2 \nabla_{\boldsymbol{\theta}}\mathcal{L}(\boldsymbol{\theta}; \mathcal{G})}_{\textnormal{Gradient\ from\ }g_{\boldsymbol{\theta}}}  - \underbrace{\beta_2 \nabla_{\boldsymbol{\theta}} \mathcal{L}_{\text{GNN}} (\boldsymbol{W}, \boldsymbol{\theta}; \mathcal{G})}_{\textnormal{Gradient\ from\ }f_{\mathbf{W}}} \label{eq: update theta}
\end{align}
\end{small}
where $\beta_1$ and $\beta_2$ are two learning rates for updating $\boldsymbol{W}$ and $\boldsymbol{\theta}$, respectively.
See Algorithm~\ref{alg: joint train} for a full description.
A baseline named ``Pre-trained'' optimizes $\boldsymbol{\theta}$ and $\boldsymbol{W}$ separately.
After optimizing $g_{\boldsymbol{\theta}}$ by minimizing $\mathcal{L}(\boldsymbol{\theta}; \mathcal{G})$ in Eq.~(\ref{eq: cross-entropy loss for g}),
$f_{\boldsymbol{W}}$ will treat the inferred $\hat{A}'$ as a fixed constant feature and minimize $\mathcal{L}_{\text{GNN}} (\boldsymbol{W}; \mathcal{G})$ in Eq.~(\ref{eq: fairGNN loss}).
\vspace{-0.2cm}
\begin{algorithm}
\caption{Joint training for $g_{\boldsymbol{\theta}}$ and $f_{\boldsymbol{W}}$.}
\label{alg: joint train}
\begin{algorithmic}
\small
\State \textbf{Input}: graph $\mathcal{G}$; node features $\boldsymbol{X}$; sensitive attribute $A$; training epochs $T$; hyper-parameter $\lambda$, learning rates $\beta_1$ and $\beta_2$, and replication times $k$.
\State \textbf{Output}: optimal $\boldsymbol{\theta}$ and $\boldsymbol{W}$ of $g_{\boldsymbol{\theta}}$ and $f_{\boldsymbol{W}}$, respectively.
\State Initialize parameters $\boldsymbol{\theta}$ and $\boldsymbol{W}$.
\State Replicate users and reviews for $k$ times as in Section~\ref{sec:minor_user_augmentation}. \Comment{\small\textcolor{blue}{Augmentation for $\mathcal{V}^U_{0, 1}$.}}
\For{$t=1,\dots, T$}
    \State Prune non-spam edges as in Section~\ref{sec:minor_user_augmentation}.
    \Comment{\small{\textcolor{blue}{Add data variations}}}
    \State Infer $\pr(A'=1)$ for users using $g_{\boldsymbol{\theta}}$. 
    \State Concatenate $\hat{A}'$ to user feature vectors as in Section~\ref{sec 3.3: joint training}.
    \State Mixup between two reviews sampled from $S_{0}^{\text{Tr}} $ as in Eq.~(\ref{eq: mixup first node}) and one of $\left\{S_{1}^{\text{Tr}}, S_{0}^{\text{Te}}, S_{1}^{\text{Te}}\right\}$ as in Eq.~(\ref{eq: mixup second node}). \Comment{\small{\textcolor{blue}{Augmentation for $\mathcal{V}^R_0$.}}}
    \State Update $\boldsymbol{W}$ and $\boldsymbol{\theta}$ following Eq.~(\ref{eq: update W}) and (\ref{eq: update theta}).
\EndFor
\end{algorithmic}
\end{algorithm}
\vspace{-0.5cm}
\section{EXPERIMENTS}
We seek to answer the following research questions: \\
\textbf{RQ1}: Do fairness issues exist between the ``favored'' and the ``protected'' groups, and between the ``mixed'' and the ``pure'' groups when using a GNN spam detector? 
Will the inferred $A'$ improve the group fairness? \\
\textbf{RQ2}: How to infer the subgroup memberships $A'$ defined in Eq.~(\ref{eq. definition for A'}) with a shortage of training examples?\\
 \textbf{RQ3}: Can the joint training method simultaneously promote the AUC of predicted $A'$ and group fairness?\\
 \textbf{RQ4}: Does the accuracy of predicting $A'$ contribute to the improvement in group fairness?
\subsection{Experimental Settings}
\textbf{Datasets.} We used three Yelp review datasets (``Chi'', ``NYC'', and ``Zip'' for short, see Table~\ref{tab:dataset}) which are commonly used in previous spam detection tasks~\cite{burkholder2021certification, dou2020enhancing}.
For the cutoff degree of user nodes ($p$-th percentile in Section~\ref{subsec: fairness group}), we treat $p\in \{20, 15, 10\}$ as a hyper-parameter distinguishing favored (top $p\%$ high-degree users, $A=0$) from protected groups (the remaining users, $A=1$). 
Reviews have the same value of $A$ as their associated users.
The favored users are further split into pure ($A'=0$) and mixed ($A'=1$) users following Eq.~(\ref{eq. definition for A'}).
Users are divided into training ($30\%$), validation ($10\%$), and test ($60\%$) sets with their associated reviews.
\vspace{-0cm}
\begin{table}[t]
    \centering
    \scriptsize
    \caption{Statistics of the datasets. We list the numbers of products, reviews, and users with the proportion of favored users/reviews ($\%\mathcal{V}_0^U$/$\%\mathcal{V}_0^R$) and mixed users/reviews ($\%\mathcal{V}_{0, 1}^U$/$\%\mathcal{V}_{0, 1}^R$) under 20-th, 15-th, and 10-th percentile (PC) of cutoff degree of the user groups. 
    The last column gives the ratio of spam in the group of $A=0$ to $A=1$.
    }
    \setlength{\tabcolsep}{6pt}{
    \begin{tabular}{l c c c c c c c}
    \toprule
    \specialrule{0em}{2pt}{2pt}
        \multirow{2}*{\textbf{Name}} & \multicolumn{6}{c}{\textbf{Data Statistics}} &\multirow{2}*{ $\frac{P(Y=1|A=0)}{P(Y=1|A=1)}$} \\
        \cline{2-7}
        \specialrule{0em}{2pt}{2pt}
        ~ & $|\mathcal{V}^P|$ & \multicolumn{3}{c}{$|\mathcal{V}^R|$} &  \multicolumn{2}{c}{$|\mathcal{V}^U|$} & \\
        \cline{1-8}
        \specialrule{0em}{2pt}{2pt}
        \multirow{5}*{\textbf{Chi}} & \multirow{5}*{$201$} & \multicolumn{3}{c}{$67,395$} & \multicolumn{2}{c}{$38,063$} & \\
        \cline{3-7}
        \specialrule{0em}{1pt}{1pt}
        ~ & ~ & \textbf{PC}& $\%\mathcal{V}^R_0$ & $\%\mathcal{V}^R_{0,1}$ & $\%\mathcal{V}^U_0$ & $\%\mathcal{V}^U_{0,1}$  & ~ \\
        \cline{3-7}
        \specialrule{0em}{1pt}{1pt}
        ~ & ~ & 20th & $13.581\%$ & $0.116\%$ & $1.781\%$ & $0.011\%$ & $0.0438$ \\
        ~ & ~ & 15th & $18.687\%$ & $0.129\%$ & $3.003\%$ & $0.026\%$ & $0.0125$ \\
        ~ & ~ & 10th & $27.003\%$ & $0.224\%$ & $5.735\%$ & $0.058\%$ & $0.0282$ \\
        \cline{1-8}
        \specialrule{0em}{2pt}{2pt}
        \multirow{5}*{\textbf{NYC}} & \multirow{5}*{$923$} & \multicolumn{3}{c}{$358,911$} & \multicolumn{2}{c}{$160,220$} & \\
        \cline{3-7}
        \specialrule{0em}{1pt}{1pt}
        ~ & ~ & \textbf{PC} & $\%\mathcal{V}^R_0$ & $\%\mathcal{V}^R_{0,1}$ & $\%\mathcal{V}^U_0$ & $\%\mathcal{V}^U_{0,1}$  & ~ \\
        \cline{3-7}
        \specialrule{0em}{1pt}{1pt}
        ~ & ~ & 20th & $12.665\%$ & $0.193\%$ & $0.760\%$ & $0.009\%$ & $0.0479$ \\
        ~ & ~ & 15th & $16.264\%$ & $0.258\%$ & $1.171\%$ & $0.018\%$ & $0.0289$ \\
        ~ & ~ & 10th & $23.096\%$ & $0.360\%$ & $2.260\%$ & $0.034\%$ & $0.0420$ \\
        \cline{1-8}
        \specialrule{0em}{2pt}{2pt}
        \multirow{5}*{\textbf{Zip}} & \multirow{5}*{$5,044$} & \multicolumn{3}{c}{$608,598$} & \multicolumn{2}{c}{$260,277$} & \\
        \cline{3-7}
        \specialrule{0em}{1pt}{1pt}
        ~ & ~ & \textbf{PC}& $\%\mathcal{V}^R_0$ & $\%\mathcal{V}^R_{0,1}$ & $\%\mathcal{V}^U_0$ & $\%\mathcal{V}^U_{0,1}$  & ~ \\
        \cline{3-7}
        \specialrule{0em}{1pt}{1pt}
        ~ & ~ & 20th & $6.859\%$ & $0.050\%$ & $0.272\%$ & $0.002\%$ & $0.0426$ \\
        ~ & ~ & 15th & $15.658\%$ & $0.145\%$ & $1.020\%$ & $0.009\%$ & $0.0139$ \\
        ~ & ~ & 10th & $22.342\%$ & $0.278\%$ & $1.968\%$ & $0.018\%$ & $0.0241$ \\
    \bottomrule
    \end{tabular}}
    \label{tab:dataset}
    \vspace{-0.5cm}
\end{table}

\noindent\textbf{Evaluation Metrics.}
(1) NDCG is to evaluate the detector's accuracy, where a higher NDCG means that the detector tends to assign a higher suspiciousness to spams than non-spams.
(2) To evaluate the ranking of spams within the group of $A=0$, a metric called ``\textbf{Av}era\textbf{g}e \textbf{F}alse \textbf{R}anking \textbf{R}atio'' is proposed to measure the average of relative ranking between spams from mixed and pure users
\begin{small}
\begin{equation}\label{eq: afrr}
\setlength{\abovedisplayskip}{5pt}
\setlength{\belowdisplayskip}{5pt}
    \text{AFRR}_{A'} = \frac{1}{Z} \sum_{y_j =1\atop v_j\in \mathcal{V}^{R}_{0, A'}} \frac{\sum_{i=1}^{|\mathcal{V}^R_{0}|} \mathbbm{1}\left[\hat{y}_i > \hat{y}_j, y_i=0\right]}{\sum_{i=1}^{|\mathcal{V}^R_0|} \mathbbm{1}[y_i =0]}, \quad A' \in \{0, 1\}
\end{equation}
\end{small}
\noindent where $A' \in \{0, 1\}$ denotes the subgroup membership.
$Z$ is the number of spams from a subgroup.
$\mathcal{V}^{R}_{0, A'}$ denotes the reviews from a subgroup users.
The ratio in the above equation calculates the proportion of non-spams ranked higher than spams over all the non-spams from the group of $A=0$. 
The lower the AFRR, the fewer non-spams ranked higher than spams.
Compared to NDCG, AFRR considers the non-spams across different subgroups and ignores the relative ranking of spams from the other subgroup.
(3) AUC is to evaluate the performance of $g_{\bt}$ in predicting $A'$.
The larger the AUC value, the more accurate $A'$ given by $g_{\bt}$.

\noindent\textbf{Methods compared.} \textbf{Joint+GNN-}$\mathbf{S_{1}^{\textbf{Tr}}}$ denotes our method: \textbf{Joint} is the joint training for two GNNs ($f_{\bw}$ and $g_{\bt}$ in Section~\ref{sec 3.3: joint training}),
\textbf{GNN-}$\mathbf{S_{1}^{\textbf{Tr}}}$ is a GNN with our mixup method in Eq.~(\ref{eq: mixup second node}).
``a\textbf{+}b'' denotes a setting: ``a'' is one of a method for obtaining the value of $A'$; ``b'' is a spam detector.

\noindent\textbf{Baselines for obtaining the value of $A'$ (selections for ``a''):}
\begin{itemize}[leftmargin=*]
    \item \textbf{w/o}: has no definition of $A'$.
    \item \textbf{Random}: randomly assigns $1/0$ to $A'$.
    \item \textbf{GT}: assigns the ground truth of $A'$ for users, which is the ideal case, as $A^\prime$ is unknown.
    \item \textbf{Pre-trained}: is a variant of \text{Joint} that pre-training $g_{\boldsymbol{\theta}}$ to infer $A'$ and fixes the inferred $A'$ when training $f_{\boldsymbol{W}}$. 
\end{itemize}
\noindent \textbf{Baselines for the spam detectors (selections for ``b''):}
\begin{itemize}[leftmargin=*]
    \item \textbf{FairGNN}~\cite{dai2021say}: is an adversarial method to get fair predictions for all groups defined by the known $A$. 
    \item \textbf{EDITS}~\cite{dong2022edits}: modifies the node attribute and the graph structure to debias the graph. 
    FairGNN and EDITS consider $A$ as a known sensitive attribute and exclude any information about $A'$ defined by our work.
    \item \textbf{GNN}: is the vanilla GNN.
    \item \textbf{GNN-}$\mathbf{S_{0}^{\textbf{Te}}}$ is a GNN with the mixup Case 2 in Eq.~(\ref{eq: mixup second node}).
    \item \textbf{GNN-}$\mathbf{S_{1}^{\textbf{Te}}}$ is a GNN with the mixup Case 3 in Eq.~(\ref{eq: mixup second node}).
\end{itemize}
We set $T=300$, $\lambda=5$, $\beta_1=\beta_2=0.001$, weight decay $=0.0001$ for both $f_{\boldsymbol{W}}$ and $g_{\boldsymbol{\theta}}$ in Algorithm~\ref{alg: joint train}, mixup weight $\alpha=0.8$.
We have 10 training-validation-test splits of the three datasets. 
The following results are based on the aggregated performance on all splits.
\begin{table*}
\vspace{-0.2cm}
    \centering
    \footnotesize
    \setlength{\abovecaptionskip}{-0.01cm}
    \caption{NDCG values for the outputs of spam detectors on three Yelp datasets with \textcolor{blue}{$20$-percentile} of user degree as a cutoff for group of $A$.
    Shown are the mean and standard deviation of $\Delta_{\text{NDCG}}$ over all 10 splits.
    ``\textbf{\textcolor{red}{*}}'' denotes our method. 
    The lower the $\Delta_{\text{NDCG}}$, the fairer the model.} 
    \resizebox{\textwidth}{!}{
    \begin{tabular}{c|c|c c c | c c c |c c c}
    \toprule
        \textbf{Detector} & \textbf{Metrics(\%)} & \multicolumn{3}{c|}{\textbf{Chi}} & \multicolumn{3}{c|}{\textbf{NYC}} & \multicolumn{3}{c}{\textbf{Zip}} \\  
         \cline{1-11}
        \multirow{3}*{\textbf{FairGNN}~\cite{dai2021say}} & NDCG($\mathcal{V}^R$) & \multicolumn{3}{c|}{86.2\tiny{$\pm $0.1}} & \multicolumn{3}{c|}{85.9\tiny{$\pm $0.0}}  & \multicolumn{3}{c}{89.6\tiny{$\pm $0.5}} \\
        ~ & NDCG($\mathcal{V}^R_1$) & \multicolumn{3}{c|}{86.2\tiny{$\pm $0.1}} & \multicolumn{3}{c|}{86.0\tiny{$\pm $0.0}}  & \multicolumn{3}{c}{89.6\tiny{$\pm $0.4}} \\
        \rowcolor{gray!20}
         \cellcolor{white} ~ & $\Delta_{\text{NDCG}}$ & \multicolumn{3}{c|}{53.7\tiny{$\pm $2.1}} & \multicolumn{3}{c|}{20.7\tiny{$\pm $1.1}} & \multicolumn{3}{c}{36.4\tiny{$\pm $4.6}} \\
        \cline{1-11}
         \multirow{3}*{\textbf{EDITS}~\cite{dong2022edits}} & NDCG($\mathcal{V}^R$) & \multicolumn{3}{c|}{84.3\tiny{$\pm $0.3}} & \multicolumn{3}{c|}{84.9\tiny{$\pm $0.1}} & \multicolumn{3}{c}{89.2\tiny{$\pm $0.0}} \\
          ~ & NDCG($\mathcal{V}^R_1$) & \multicolumn{3}{c|}{84.3\tiny{$\pm $0.3}} & \multicolumn{3}{c|}{85.1\tiny{$\pm $0.1}} & \multicolumn{3}{c}{89.3\tiny{$\pm $0.0}}\\
        \rowcolor{gray!20} 
         \cellcolor{white} ~ & $\Delta_{\text{NDCG}}$ & \multicolumn{3}{c|}{50.9\tiny{$\pm $2.3}} & \multicolumn{3}{c|}{32.7\tiny{$\pm $1.5}} & \multicolumn{3}{c}{39.4\tiny{$\pm $10.6}} \\
        \cline{1-11}
        \midrule
          \multirow{2}*{\textbf{Detector}}&  \multirow{2}*{\textbf{Metrics(\%)}}  & \multicolumn{3}{c|}{\textbf{GNN}$(g_{\boldsymbol{\theta}})$} & \multicolumn{3}{c|}{\textbf{GNN}$(g_{\boldsymbol{\theta}})$} & \multicolumn{3}{c}{\textbf{GNN}$(g_{\boldsymbol{\theta}})$}\\
         &   ~&  \textbf{w/o} & \textbf{Pre-trained} &  \textbf{Joint\textcolor{red}{*}}& \textbf{w/o} & \textbf{Pre-trained} &  \textbf{Joint\textcolor{red}{*}} & \textbf{w/o} & \textbf{Pre-trained} &  \textbf{Joint\textcolor{red}{*}}\\
         \cline{1-11}
     \multirow{3}*{\textbf{GNN}} &  NDCG($\mathcal{V}^R$) & 84.5\tiny{$\pm$0.9}  & 83.2\tiny{$\pm $1.5} & 83.3\tiny{$\pm $2.2}  & 85.2\tiny{$\pm $0.8} & 85.1\tiny{$\pm$0.8} &  85.2\tiny{$\pm$0.5} & 88.4\tiny{$\pm$1.4} & 87.6\tiny{$\pm$1.5} & 88.6\tiny{$\pm$1.0}\\    
        ~ & NDCG($\mathcal{V}^R_1$) & 84.7\tiny{$\pm $0.9} & 83.5\tiny{$\pm $1.3} & 83.6\tiny{$\pm $2.1} & 85.1\tiny{$\pm$ 0.8} & 85.3\tiny{$\pm$0.8} & 85.2\tiny{$\pm$0.5} & 88.4\tiny{$\pm$1.4} & 87.6\tiny{$\pm$1.5} & 88.6\tiny{$\pm$1.0}\\
      \rowcolor{gray!20} 
     \cellcolor{white} ~ & $\Delta_{\text{NDCG}}$ & 51.2\tiny{$\pm $2.1} & 51.0\tiny{$\pm $3.0} & \textbf{50.7}\tiny{$\pm $3.5}  & 21.9\tiny{$\pm$7.0} & 21.8\tiny{$\pm$7.8} & \textbf{21.3}\tiny{$\pm$8.9} &  36.3\tiny{$\pm$10.4} & \textbf{34.3}\tiny{$\pm$10.8} &  34.8\tiny{$\pm$11.1}\\     
     \cline{1-11}
         \multirow{3}*{\textbf{GNN-}$\mathbf{S_{1}^{\textbf{Tr}}}$\textbf{\textcolor{red}{*}}} &  NDCG($\mathcal{V}^R$) & 85.6\tiny{$\pm $0.7} & 85.8\tiny{$\pm $0.5} & 85.6\tiny{$\pm $0.8}  & 85.8\tiny{$\pm$0.1} & 85.9\tiny{$\pm$0.0} &  85.9\tiny{$\pm$0.0} &  89.7\tiny{$\pm$0.2} & 89.7\tiny{$\pm$0.1} & 89.6\tiny{$\pm$0.1}\\
      ~  & NDCG($\mathcal{V}^R_1$) & 85.6\tiny{$\pm $0.7} & 85.8\tiny{$\pm $0.5} & 85.7\tiny{$\pm $0.8}  & 85.9\tiny{$\pm$0.0} & 86.0\tiny{$\pm$0.0} & 86.0\tiny{$\pm$0.0} & 89.7\tiny{$\pm$0.2} & 89.7\tiny{$\pm$0.1} & 89.6\tiny{$\pm$0.1}\\
       \rowcolor{gray!20}
     \cellcolor{white}  ~ & $\Delta_{\text{NDCG}}$ & 51.6\tiny{$\pm $0.9} & 50.3\tiny{$\pm $1.0} & \textbf{50.1}\tiny{$\pm $1.0}  & 19.1\tiny{$\pm$5.5} & 19.0\tiny{$\pm$5.2} & \textbf{17.9}\tiny{$\pm$6.1} & 38.7\tiny{$\pm$7.2} & 36.0\tiny{$\pm$9.0} & \textbf{34.3}\tiny{$\pm$11.5}\\ 
 \cline{1-11}
       \multirow{3}*{\textbf{GNN-}$\mathbf{S_{0}^{\textbf{Te}}}$} &  NDCG($\mathcal{V}^R$) & 85.1\tiny{$\pm $0.7} & 85.3\tiny{$\pm $1.4} & 85.2\tiny{$\pm $1.4} & 85.3\tiny{$\pm$0.6} & 85.4\tiny{$\pm$0.5} & 85.4\tiny{$\pm$0.4} &  89.4\tiny{$\pm$0.6} & 89.6\tiny{$\pm$0.1} & 89.0\tiny{$\pm$0.06} \\
       ~ & NDCG($\mathcal{V}^R_1$) & 85.2\tiny{$\pm $0.8} & 83.4\tiny{$\pm $1.3} & 83.5\tiny{$\pm $1.3} & 85.2\tiny{$\pm$0.6} & 85.5\tiny{$\pm$0.4} & 85.5\tiny{$\pm$0.4} & 89.4\tiny{$\pm$0.06} & 89.0\tiny{$\pm$0.01} & 89.1\tiny{$\pm$0.06}\\
       \rowcolor{gray!20}
       \cellcolor{white} ~ & $\Delta_{\text{NDCG}}$ & 51.2\tiny{$\pm $1.5} & 
       \textbf{50.9}\tiny{$\pm $2.4} & \textbf{50.9}\tiny{$\pm $2.3}  & 21.9\tiny{$\pm$6.9}  & 21.9\tiny{$\pm$6.3}  & \textbf{20.9}\tiny{$\pm$9.4} & 38.9\tiny{$\pm$7.6} & 36.9\tiny{$\pm$9.5} & \textbf{34.9}\tiny{$\pm$11.0}\\   
 \cline{1-11}
        \multirow{3}*{\textbf{GNN-}$\mathbf{S_{1}^{\textbf{Te}}}$} &  NDCG($\mathcal{V}^R$) & 84.7\tiny{$\pm $1.3} & 83.7\tiny{$\pm $0.9} & 83.1\tiny{$\pm $0.9}  & 85.7\tiny{$\pm$0.1} & 85.8\tiny{$\pm$0.1} & 85.8\tiny{$\pm$0.2} &  89.6\tiny{$\pm$0.3} & 89.6\tiny{$\pm$0.1} & 89.5\tiny{$\pm$0.3} \\
       ~ & NDCG($\mathcal{V}^R_1$) & 84.8\tiny{$\pm $1.3} & 83.9\tiny{$\pm $0.9} & 83.4\tiny{$\pm $0.9} & 85.8\tiny{$\pm$0.1} & 85.8\tiny{$\pm$0.1} & 85.8\tiny{$\pm$0.1} & 89.6\tiny{$\pm$0.3} & 89.6\tiny{$\pm$0.1} & 89.5\tiny{$\pm$0.3}\\
       \rowcolor{gray!20}
       \cellcolor{white} ~ & $\Delta_{\text{NDCG}}$ & 51.3\tiny{$\pm $0.6} & 50.8\tiny{$\pm $0.6} & \textbf{50.2}\tiny{$\pm $1.0}  &  21.0\tiny{$\pm$5.4}  & 19.8\tiny{$\pm$5.4}  & \textbf{19.3}\tiny{$\pm$6.8} & 38.7\tiny{$\pm$7.2} & 36.2\tiny{$\pm$9.5} & \textbf{34.6}\tiny{$\pm$11.4}\\   
    \bottomrule
    \end{tabular}}
    \label{tab: big table 20-percentile}
    \vspace{-0.2cm}
\end{table*}
\begin{table*}
    \centering
    \footnotesize
    \setlength{\abovecaptionskip}{-0.01cm}
    \caption{NDCG values for the outputs of detectors on Yelp datasets with \textcolor{blue}{$15$-percentile} of user degree as a cutoff for group of $A$.}
    \resizebox{\textwidth}{!}{
    \begin{tabular}{c|c|c c c | c c c |c c c}
    \toprule
        \textbf{Detector} & \textbf{Metrics(\%)} & \multicolumn{3}{c|}{\textbf{Chi}} & \multicolumn{3}{c|}{\textbf{NYC}} & \multicolumn{3}{c}{\textbf{Zip}} \\  
         \cline{1-11}
        \multirow{3}*{\textbf{FairGNN}~\cite{dai2021say}} & NDCG($\mathcal{V}^R$) & \multicolumn{3}{c|}{86.2\tiny{$\pm $0.2}} & \multicolumn{3}{c|}{85.9\tiny{$\pm $0.1}}  & \multicolumn{3}{c}{89.9\tiny{$\pm $0.0}} \\ 
        ~ & NDCG($\mathcal{V}^R_1$) & \multicolumn{3}{c|}{86.3\tiny{$\pm $0.2}} & \multicolumn{3}{c|}{86.0\tiny{$\pm $0.1}}  & \multicolumn{3}{c}{90.0\tiny{$\pm $0.0}} \\ 
        \rowcolor{gray!20}
         \cellcolor{white} ~ & $\Delta_{\text{NDCG}}$ & \multicolumn{3}{c|}{45.8\tiny{$\pm $1.7}} & \multicolumn{3}{c|}{20.0\tiny{$\pm $2.4}} & \multicolumn{3}{c}{31.0\tiny{$\pm $2.7}} \\  
        \cline{1-11}
         \multirow{3}*{\textbf{EDITS}~\cite{dong2022edits}} & NDCG($\mathcal{V}^R$) & \multicolumn{3}{c|}{84.3\tiny{$\pm $0.2}} & \multicolumn{3}{c|}{85.0\tiny{$\pm $0.1}} & \multicolumn{3}{c}{89.2\tiny{$\pm $0.0}} \\
          ~ & NDCG($\mathcal{V}^R_1$) & \multicolumn{3}{c|}{84.4\tiny{$\pm $0.2}} & \multicolumn{3}{c|}{85.1\tiny{$\pm $0.1}} & \multicolumn{3}{c}{89.3\tiny{$\pm $0.0}}\\
        \rowcolor{gray!20} 
         \cellcolor{white} ~ & $\Delta_{\text{NDCG}}$ & \multicolumn{3}{c|}{43.8\tiny{$\pm $2.6}} & \multicolumn{3}{c|}{32.4\tiny{$\pm $1.5}} & \multicolumn{3}{c}{34.4\tiny{$\pm $3.5}} \\ 
        \cline{1-11}
        \midrule
          \multirow{2}*{\textbf{Detector}}&  \multirow{2}*{\textbf{Metrics(\%)}}  & \multicolumn{3}{c|}{\textbf{GNN}$(g_{\boldsymbol{\theta}})$} & \multicolumn{3}{c|}{\textbf{GNN}$(g_{\boldsymbol{\theta}})$} & \multicolumn{3}{c}{\textbf{GNN}$(g_{\boldsymbol{\theta}})$}\\
         &   ~&  \textbf{w/o} & \textbf{Pre-trained} &  \textbf{Joint\textcolor{red}{*}}& \textbf{w/o} & \textbf{Pre-trained} &  \textbf{Joint\textcolor{red}{*}} & \textbf{w/o} & \textbf{Pre-trained} &  \textbf{Joint\textcolor{red}{*}}\\
         \cline{1-11}
     \multirow{3}*{\textbf{GNN}} &  NDCG($\mathcal{V}^R$) & 84.3\tiny{$\pm$1.3}  & 84.3\tiny{$\pm$0.9} & 84.4\tiny{$\pm$0.9}  & 85.7\tiny{$\pm $0.2} & 84.5\tiny{$\pm$0.4} &  84.6\tiny{$\pm$0.5} & 89.5\tiny{$\pm$0.2} & 88.9\tiny{$\pm$0.6} & 88.9\tiny{$\pm$0.6}\\       
        ~ & NDCG($\mathcal{V}^R_1$) & 84.6\tiny{$\pm $1.1} & 84.7\tiny{$\pm $0.7} & 84.8\tiny{$\pm $0.7} & 85.8\tiny{$\pm$ 0.2} & 84.6\tiny{$\pm$0.4} & 84.6\tiny{$\pm$0.5} & 89.5\tiny{$\pm$0.2} & 88.9\tiny{$\pm$0.6} & 88.9\tiny{$\pm$0.7}\\  
      \rowcolor{gray!20} 
     \cellcolor{white} ~ & $\Delta_{\text{NDCG}}$ & 41.8\tiny{$\pm $2.4} & \textbf{40.9}\tiny{$\pm $3.7} & \textbf{40.9}\tiny{$\pm $4.0}  & 16.3\tiny{$\pm$3.9} & 15.8\tiny{$\pm$2.7} & \textbf{15.2}\tiny{$\pm$6.1} &  26.1\tiny{$\pm$3.4} & 26.6\tiny{$\pm$2.5} &  \textbf{25.9}\tiny{$\pm$2.7}\\     
     \cline{1-11}
         \multirow{3}*{\textbf{GNN-}$\mathbf{S_{1}^{\textbf{Tr}}}$\textbf{\textcolor{red}{*}}} &  NDCG($\mathcal{V}^R$) & 86.0\tiny{$\pm $0.4} & 85.9\tiny{$\pm $0.3} & 85.9\tiny{$\pm $0.2}  & 85.9\tiny{$\pm$0.1} & 85.9\tiny{$\pm$0.1} &  85.9\tiny{$\pm$0.1} &  89.7\tiny{$\pm$0.1} & 89.9\tiny{$\pm$0.0} & 87.9\tiny{$\pm$2.1}\\  
      ~  & NDCG($\mathcal{V}^R_1$) & 86.0\tiny{$\pm $0.4} & 86.0\tiny{$\pm $0.3} & 86.0\tiny{$\pm $0.2}  & 85.9\tiny{$\pm$0.1} & 85.9\tiny{$\pm$0.1} & 85.9\tiny{$\pm$0.1} & 89.8\tiny{$\pm$0.1} & 89.9\tiny{$\pm$0.0} & 87.9\tiny{$\pm$2.1}\\   
       \rowcolor{gray!20}
     \cellcolor{white}  ~ & $\Delta_{\text{NDCG}}$ & 43.2\tiny{$\pm $2.0} & 41.9\tiny{$\pm $5.8} & \textbf{40.5}\tiny{$\pm $5.1}  & 16.5\tiny{$\pm$3.8} & \textbf{15.6}\tiny{$\pm$2.9} & \textbf{15.6}\tiny{$\pm$3.6} & 26.4\tiny{$\pm$3.3} & 27.0\tiny{$\pm$3.2} & \textbf{22.7}\tiny{$\pm$4.2}\\ 
 \cline{1-11}
        \multirow{3}*{\textbf{GNN-}$\mathbf{S_{0}^{\textbf{Te}}}$} &  NDCG($\mathcal{V}^R$) & 84.4\tiny{$\pm $1.5} & 84.3\tiny{$\pm $1.1} & 84.4\tiny{$\pm $0.9}  & 85.8\tiny{$\pm$0.2} & 84.7\tiny{$\pm$0.4} & 84.7\tiny{$\pm$0.4} &  89.4\tiny{$\pm$0.2} & 89.1\tiny{$\pm$0.6} & 89.1\tiny{$\pm$0.6} \\ 
       ~ & NDCG($\mathcal{V}^R_1$) & 84.7\tiny{$\pm $1.1} & 84.6\tiny{$\pm $0.9} & 84.8\tiny{$\pm $0.} & 85.9\tiny{$\pm$0.2} & 84.7\tiny{$\pm$0.5} & 84.7\tiny{$\pm$0.5} & 89.5\tiny{$\pm$0.2} & 89.1\tiny{$\pm$0.6} & 89.2\tiny{$\pm$0.6}\\ 
       \rowcolor{gray!20}
       \cellcolor{white} ~ & $\Delta_{\text{NDCG}}$ & 43.4\tiny{$\pm $2.4} & 
       41.7\tiny{$\pm $4.6} & \textbf{41.1}\tiny{$\pm $4.2}  & 16.6\tiny{$\pm$4.1}  & 15.7\tiny{$\pm$3.0}  & \textbf{15.6}\tiny{$\pm$3.6} & 26.2\tiny{$\pm$4.0} & 27.2\tiny{$\pm$3.2} & \textbf{25.3}\tiny{$\pm$2.7}\\   
 \cline{1-11}
       \multirow{3}*{\textbf{GNN-}$\mathbf{S_{1}^{\textbf{Te}}}$} &  NDCG($\mathcal{V}^R$) & 85.9\tiny{$\pm $0.5} & 85.8\tiny{$\pm $0.3} & 85.9\tiny{$\pm $0.3} & 85.8\tiny{$\pm$0.1} & 85.4\tiny{$\pm$0.3} & 85.5\tiny{$\pm$0.3} &  89.8\tiny{$\pm$0.1} & 89.9\tiny{$\pm$0.1} & 89.9\tiny{$\pm$0.1} \\  
       ~ & NDCG($\mathcal{V}^R_1$) & 85.9\tiny{$\pm $0.5} & 85.9\tiny{$\pm $0.3} & 85.9\tiny{$\pm $0.3} & 85.9\tiny{$\pm$0.2} & 85.5\tiny{$\pm$0.3} & 85.5\tiny{$\pm$0.3} & 89.8\tiny{$\pm$0.1} & 89.9\tiny{$\pm$0.1} & 89.9\tiny{$\pm$0.1}\\ 
       \rowcolor{gray!20}
       \cellcolor{white} ~ & $\Delta_{\text{NDCG}}$ & 42.6\tiny{$\pm $2.8} & \textbf{40.7}\tiny{$\pm $3.7} & \textbf{40.7}\tiny{$\pm $3.7}  & 16.0\tiny{$\pm$3.9}  & \textbf{15.5}\tiny{$\pm$3.0}  & 15.6\tiny{$\pm$3.3} & 26.1\tiny{$\pm$3.5} & 27.1\tiny{$\pm$2.4} & \textbf{24.9}\tiny{$\pm$2.7}\\   
    \bottomrule
    \end{tabular}}
    \label{tab: big table 15-percentile}
\end{table*}
\begin{table*}
    \centering
    \footnotesize
    \setlength{\abovecaptionskip}{-0.01cm}
    \caption{NDCG values for the outputs of detectors on Yelp datasets with \textcolor{blue}{$10$-percentile} of user degree as a cutoff for group of $A$.}
    \resizebox{\textwidth}{!}{
    \begin{tabular}{c|c|c c c | c c c |c c c}
    \toprule
        \textbf{Detector} & \textbf{Metrics(\%)} & \multicolumn{3}{c|}{\textbf{Chi}} & \multicolumn{3}{c|}{\textbf{NYC}} & \multicolumn{3}{c}{\textbf{Zip}} \\  
         \cline{1-11}
        \multirow{3}*{\textbf{FairGNN}~\cite{dai2021say}} & NDCG($\mathcal{V}^R$) & \multicolumn{3}{c|}{86.2\tiny{$\pm $0.2}} & \multicolumn{3}{c|}{85.9\tiny{$\pm $0.1}}  & \multicolumn{3}{c}{89.9\tiny{$\pm $0.0}} \\ 
        ~ & NDCG($\mathcal{V}^R_1$) & \multicolumn{3}{c|}{86.4\tiny{$\pm $0.2}} & \multicolumn{3}{c|}{86.0\tiny{$\pm $0.1}}  & \multicolumn{3}{c}{90.0\tiny{$\pm $0.0}} \\ 
        \rowcolor{gray!20}
         \cellcolor{white} ~ & $\Delta_{\text{NDCG}}$ & \multicolumn{3}{c|}{33.8\tiny{$\pm $5.4}} & \multicolumn{3}{c|}{22.4\tiny{$\pm $1.9}} & \multicolumn{3}{c}{24.8\tiny{$\pm $1.6}} \\  
        \cline{1-11}
         \multirow{3}*{\textbf{EDITS}~\cite{dong2022edits}} & NDCG($\mathcal{V}^R$) & \multicolumn{3}{c|}{84.3\tiny{$\pm $0.2}} & \multicolumn{3}{c|}{85.0\tiny{$\pm $0.1}} & \multicolumn{3}{c}{89.2\tiny{$\pm $0.0}} \\
          ~ & NDCG($\mathcal{V}^R_1$) & \multicolumn{3}{c|}{84.5\tiny{$\pm $0.2}} & \multicolumn{3}{c|}{85.2\tiny{$\pm $0.1}} & \multicolumn{3}{c}{89.4\tiny{$\pm $0.0}}\\
        \rowcolor{gray!20} 
         \cellcolor{white} ~ & $\Delta_{\text{NDCG}}$ & \multicolumn{3}{c|}{37.1\tiny{$\pm $2.0}} & \multicolumn{3}{c|}{30.0\tiny{$\pm $1.2}} & \multicolumn{3}{c}{28.9\tiny{$\pm $1.3}} \\ 
        \cline{1-11}
        \midrule
          \multirow{2}*{\textbf{Detector}}&  \multirow{2}*{\textbf{Metrics(\%)}}  & \multicolumn{3}{c|}{\textbf{GNN}$(g_{\boldsymbol{\theta}})$} & \multicolumn{3}{c|}{\textbf{GNN}$(g_{\boldsymbol{\theta}})$} & \multicolumn{3}{c}{\textbf{GNN}$(g_{\boldsymbol{\theta}})$}\\
         &   ~&  \textbf{w/o} & \textbf{Pre-trained} &  \textbf{Joint\textcolor{red}{*}}& \textbf{w/o} & \textbf{Pre-trained} &  \textbf{Joint\textcolor{red}{*}} & \textbf{w/o} & \textbf{Pre-trained} &  \textbf{Joint\textcolor{red}{*}}\\
         \cline{1-11}
     \multirow{3}*{\textbf{GNN}} &  NDCG($\mathcal{V}^R$) & 85.4\tiny{$\pm$0.5}  & 84.7\tiny{$\pm$1.3} & 84.9\tiny{$\pm$1.4}  & 84.8\tiny{$\pm $0.4} & 84.5\tiny{$\pm$0.3} &  84.6\tiny{$\pm$0.4} & 89.7\tiny{$\pm$0.3} & 89.6\tiny{$\pm$0.6} & 88.9\tiny{$\pm$0.6}\\       
        ~ & NDCG($\mathcal{V}^R_1$) & 85.7\tiny{$\pm $0.4} & 85.3\tiny{$\pm $0.8} & 85.4\tiny{$\pm $0.9} & 84.8\tiny{$\pm$ 0.4} & 84.6\tiny{$\pm$0.3} & 84.7\tiny{$\pm$0.4} & 89.8\tiny{$\pm$0.3} & 89.6\tiny{$\pm$0.6} & 89.8\tiny{$\pm$0.4}\\  
      \rowcolor{gray!20} 
     \cellcolor{white} ~ & $\Delta_{\text{NDCG}}$ & 34.7\tiny{$\pm $1.5} & \textbf{33.9}\tiny{$\pm $2.5} & 34.7\tiny{$\pm $1.5}  & 19.7\tiny{$\pm$1.6} & 19.1\tiny{$\pm$1.6} & \textbf{18.7}\tiny{$\pm$2.0} &  25.0\tiny{$\pm$1.8} & 23.7\tiny{$\pm$1.7} &  \textbf{23.6}\tiny{$\pm$1.7}\\     
     \cline{1-11}
         \multirow{3}*{\textbf{GNN-}$\mathbf{S_{1}^{\textbf{Tr}}}$\textbf{\textcolor{red}{*}}} &  NDCG($\mathcal{V}^R$) & 85.9\tiny{$\pm $0.4} & 86.1\tiny{$\pm $0.2} & 86.1\tiny{$\pm $0.2}  & 85.9\tiny{$\pm$0.1} & 85.9\tiny{$\pm$0.1} &  85.9\tiny{$\pm$0.1} &  89.7\tiny{$\pm$0.1} & 89.9\tiny{$\pm$0.0} & 89.9\tiny{$\pm$0.0}\\  
      ~  & NDCG($\mathcal{V}^R_1$) & 85.9\tiny{$\pm $0.4} & 86.2\tiny{$\pm $0.2} & 86.2\tiny{$\pm $0.2}  & 86.0\tiny{$\pm$0.1} & 86.0\tiny{$\pm$0.1} & 86.0\tiny{$\pm$0.1} & 89.8\tiny{$\pm$0.1} & 90.0\tiny{$\pm$0.0} & 90.0\tiny{$\pm$0.0}\\   
       \rowcolor{gray!20}
     \cellcolor{white}  ~ & $\Delta_{\text{NDCG}}$ & 34.1\tiny{$\pm $4.6} & 33.7\tiny{$\pm $3.0} & \textbf{33.2}\tiny{$\pm $2.6}  & 19.1\tiny{$\pm$1.9} & 16.8\tiny{$\pm$1.5} & \textbf{16.6}\tiny{$\pm$1.9} & 25.1\tiny{$\pm$1.6} & 23.4\tiny{$\pm$1.6} & \textbf{23.3}\tiny{$\pm$1.4}\\ 
 \cline{1-11}
       \multirow{3}*{\textbf{GNN-}$\mathbf{S_{0}^{\textbf{Te}}}$} &  NDCG($\mathcal{V}^R$) & 85.8\tiny{$\pm $0.5} & 86.1\tiny{$\pm $0.3} & 86.1\tiny{$\pm $0.2} & 85.9\tiny{$\pm$0.1} & 85.4\tiny{$\pm$0.3} & 85.4\tiny{$\pm$0.3} &  89.8\tiny{$\pm$0.1} & 90.0\tiny{$\pm$0.1} & 89.9\tiny{$\pm$0.1} \\  
       ~ & NDCG($\mathcal{V}^R_1$) & 85.9\tiny{$\pm $0.5} & 86.2\tiny{$\pm $0.3} & 86.2\tiny{$\pm $0.2} & 86.0\tiny{$\pm$0.1} & 85.5\tiny{$\pm$0.3} & 85.5\tiny{$\pm$0.3} & 89.9\tiny{$\pm$0.1} & 90.0\tiny{$\pm$0.1} & 90.0\tiny{$\pm$0.1}\\ 
       \rowcolor{gray!20}
       \cellcolor{white} ~ & $\Delta_{\text{NDCG}}$ & 33.6\tiny{$\pm $4.4} & 34.1\tiny{$\pm $3.3} & \textbf{33.8}\tiny{$\pm $3.0}  & 19.7\tiny{$\pm$2.0}  & \textbf{18.6}\tiny{$\pm$1.5}  & \textbf{18.6}\tiny{$\pm$1.9} & 25.1\tiny{$\pm$1.6} & 23.7\tiny{$\pm$1.6} & \textbf{23.6}\tiny{$\pm$1.4}\\   
 \cline{1-11}
        \multirow{3}*{\textbf{GNN-}$\mathbf{S_{1}^{\textbf{Te}}}$} &  NDCG($\mathcal{V}^R$) & 85.6\tiny{$\pm $0.5} & 85.0\tiny{$\pm $1.2} & 85.0\tiny{$\pm $1.2}  & 85.9\tiny{$\pm$0.1} & 84.7\tiny{$\pm$0.3} & 84.7\tiny{$\pm$0.3} &  89.6\tiny{$\pm$0.1} & 89.7\tiny{$\pm$0.5} & 89.5\tiny{$\pm$1.2} \\ 
       ~ & NDCG($\mathcal{V}^R_1$) & 85.7\tiny{$\pm $0.6} & 85.4\tiny{$\pm $0.8} & 85.5\tiny{$\pm $0.7} & 86.0\tiny{$\pm$0.1} & 84.8\tiny{$\pm$0.4} & 84.8\tiny{$\pm$0.4} & 89.7\tiny{$\pm$0.1} & 89.8\tiny{$\pm$0.5} & 89.2\tiny{$\pm$0.7}\\ 
       \rowcolor{gray!20}
       \cellcolor{white} ~ & $\Delta_{\text{NDCG}}$ & \textbf{34.0}\tiny{$\pm $4.6} & 34.3\tiny{$\pm $3.0} & 34.2\tiny{$\pm $2.8}  & 19.6\tiny{$\pm$1.9}  & \textbf{18.7}\tiny{$\pm$1.7}  & \textbf{18.7}\tiny{$\pm$2.1} & 25.1\tiny{$\pm$1.6} & \textbf{23.8}\tiny{$\pm$1.7} & 24.2\tiny{$\pm$2.8}\\   
    \bottomrule
    \end{tabular}}
    \label{tab: big table 10-percentile}
\end{table*}
\subsection{Results}
\subsubsection{Group Fairness.} To answer \textbf{RQ1}, we take the difference in the NDCGs of reviews from groups of $A=0$ and $A=1$ as the group fairness metric, denoted by $\Delta_{\text{NDCG}}$. 
Table~\ref{tab: big table 20-percentile}, \ref{tab: big table 15-percentile}, and \ref{tab: big table 10-percentile} show the NDCG values for the outputs of spam detectors with $20$-th, $15$-th, and $10$-th percentile of user node degree as a cutoff for groups of $A$, respectively.
Each table has rows for six detectors in two sections.
\textbf{FairGNN} and \textbf{EDITS} in the upper section ignore the attribute $A'$ that we define.
The remaining four detectors use $A'$ where the columns under each dataset column are methods for obtaining the value of $A'$. 
These tree tables report the NDCG of all test reviews NDCG($\mathcal{V}^R$), the NDCG of test reviews from the protected group NDCG($\mathcal{V}^R_{1}$), and $\Delta_{\text{NDCG}}$.

Large values of $\Delta_{\text{NDCG}}$ for \textbf{FairGNN}, \textbf{EDITS}, and \textbf{w/o+GNN} show the pervasive unfairness existing in the GNN-based models.
\textbf{FairGNN} and \textbf{EDITS} have larger $\Delta_{\text{NDCG}}$ when improving NDCGs for the favored groups, meaning they worsen the fairness: the NDCGs are increased more on the favored groups than the protected groups.
For any spam detector in the lower section, the proposed \textbf{Joint} method has the smallest $\Delta_{\text{NDCG}}$ in most cases. 
Comparing to \textbf{Pre-trained}, $g_{\bt}$ in \textbf{Joint} receives the additional gradient of $\bt$ from $f_{\bw}$ (see Eq.~(\ref{eq: update theta})). 
However, for detectors without the review and user augmentation (i.e., \textbf{Joint}+\textbf{GNN}), the additional gradient may lead $g_{\bt}$ to infer $A'$ hurting the performance of $f_{\bw}$. 
For any method obtaining the value of $A'$, the detector with our augmentation \textbf{GNN-$\text{S}_{1}^{\text{Tr}}$} has the smallest $\Delta_{\text{NDCG}}$ almost in all cases.
Since Chi and Zip contain fewer mixed users, maintaining the original distribution while mixing up is more complicated.
\begin{figure*}
\centering
\includegraphics[width=0.85\textwidth]{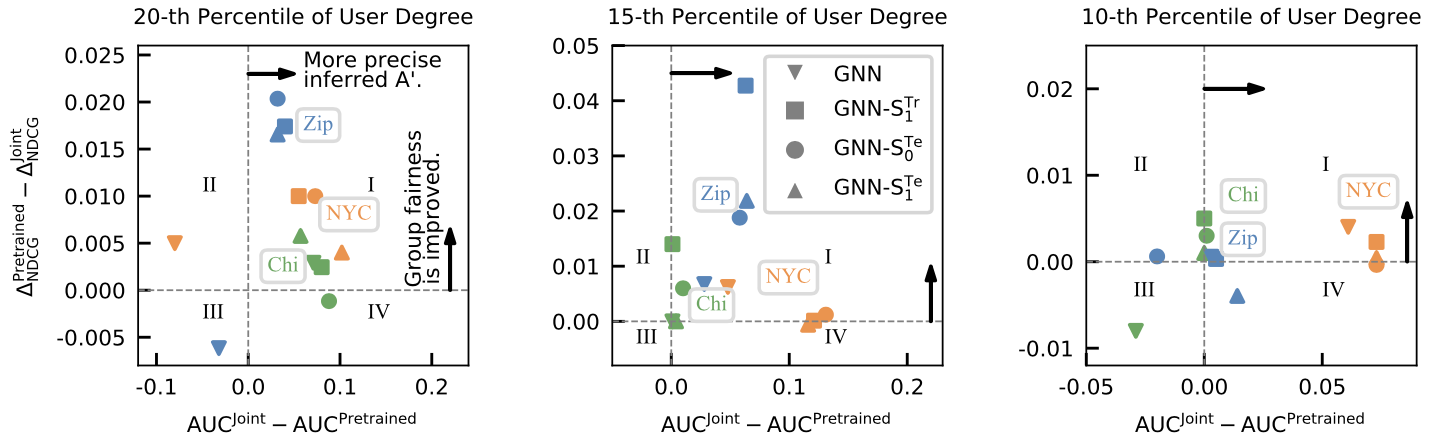}
\caption{
Relation between the accuracy of predicting $A'$ and group fairness. 
$x$-axis: the gap of $g_{\bt}$'s AUC in predicting $A'$ between Joint and Pre-trained; $y$-axis: the gap of $\Delta_{\text{NDCG}}$ of $f_{\bw}$ in spam detection between Pre-trained and Joint.
\textbf{Joint} method simultaneously promotes the AUC of predicted $A'$ and group fairness.}
\label{fig: g_theta auc}
\end{figure*}
\begin{figure*}
   \centering
   \vspace{-0.1cm}
\includegraphics[width=0.85\textwidth]{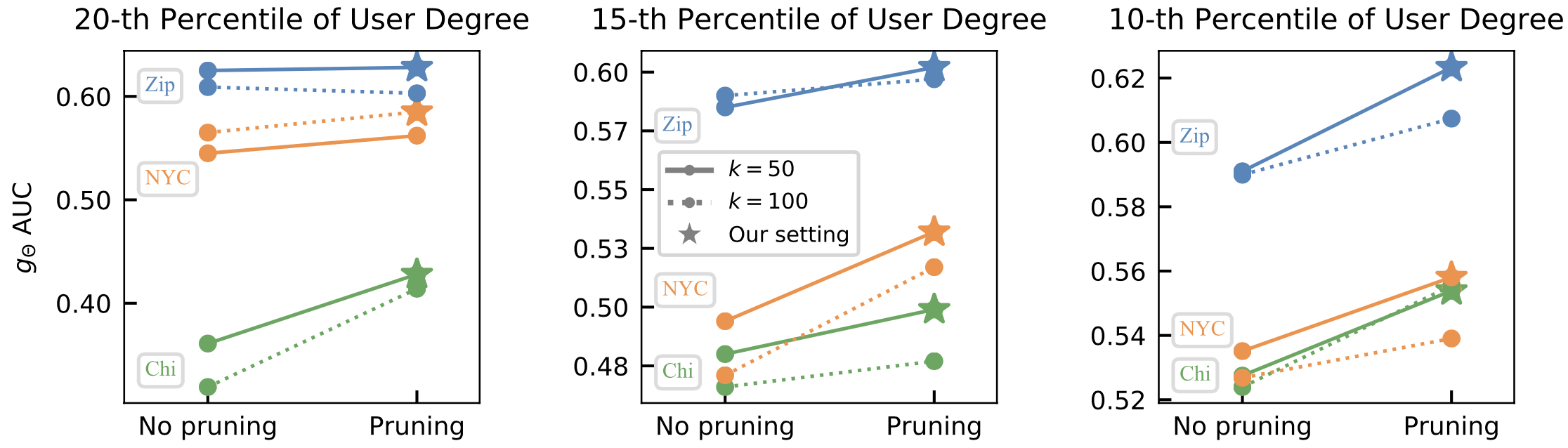}
	\caption{Sensitivity analysis for the replication time $k$ and pruning non-spam edges. This figure shows the test AUCs of $g_{\boldsymbol{\theta}}$ on graphs with various $k$ and w/ or w/o pruning edges.} 
    \label{fig: sensitive of duplication times k and pruning edges.}
    \vspace{-0.2cm}
\end{figure*}

\begin{figure*}
   \centering
   \vspace{-0.2cm}
    \subfigtopskip=0.5cm
    \subfigbottomskip = -10pt
    \subfigure[20-th Percentile of User Node Degree]{
    \includegraphics[width=0.9\textwidth]{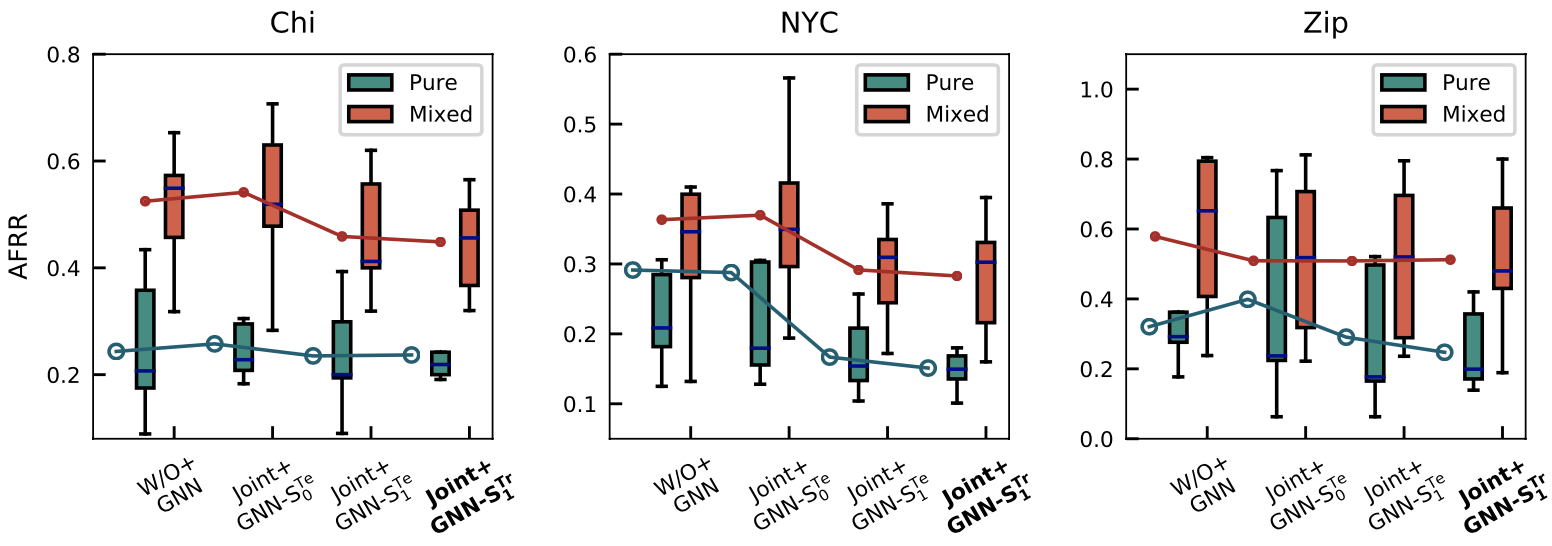}}
    \subfigbottomskip = -5pt
    \vspace{-0.2cm}
    \subfigure[15-th Percentile of User Node Degree]{
    \includegraphics[width=0.9\textwidth]{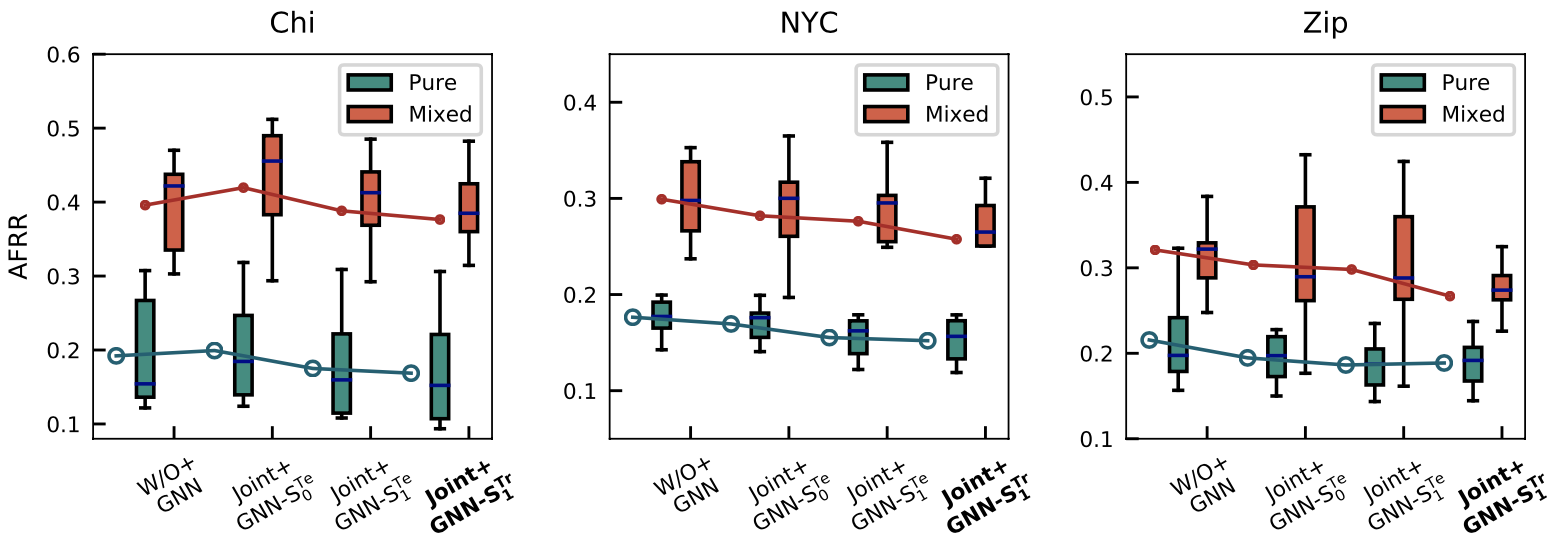}}
    \subfigbottomskip = -5pt
    \subfigure[10-th Percentile of User Node Degree]{
    \includegraphics[width=0.9\textwidth]{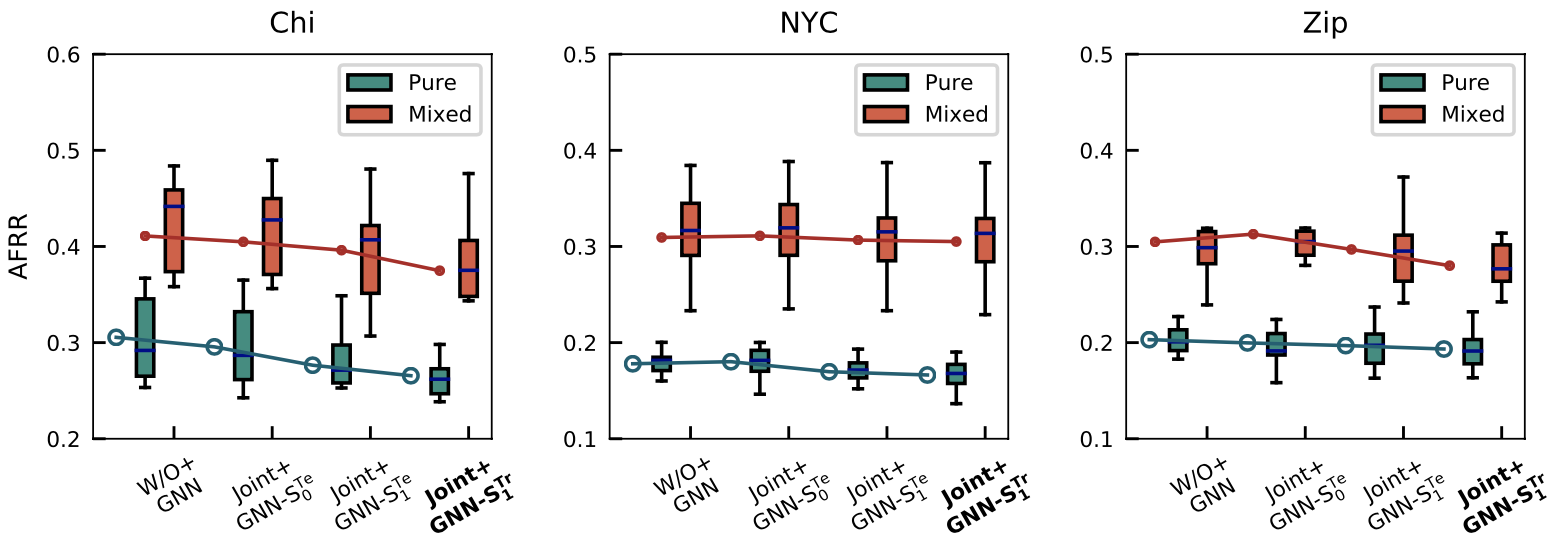}}
	\caption{Box plot of AFRR in Eq. (\ref{eq: afrr}) for test mixed and pure users of all splits of three datasets under four settings (the proposed method is bold). 
	A box records a set of AFRRs for all the splits, and the solid lines connect the mean of each box. 
	Introducing $A'$ and the joint training of $f_{\bw}$ and $g_{\bt}$ (the\textbf{Joint} method) decreases AFRR for mixed users by giving larger suspiciousness to spam than non-spams.}
    \label{fig: subgroup fairness}
\end{figure*}

\noindent\textbf{Explanation of improved group fairness.} Fig.~\ref{fig: subgroup fairness} shows the test AFRRs of pure and mixed subgroups across four methods \{\textbf{w/o+GNN}, \textbf{Joint+GNN-$\text{S}_{1}^{\text{Tr}}$}, \textbf{Joint+GNN-$\text{S}_{0}^{\text{Te}}$}, \textbf{Joint+ GNN-$\text{S}_{1}^{\text{Te}}$}\}. 
It demonstrates the effect of adding $A'$ on spams from subgroups and explains the improved NDCG on the protected group.
\textbf{w/o+GNN} shows that spams from the mixed users have larger AFRRs than the pure users for all datasets.
In other words, inside the favored group, the basic GNN tends to rank spams from pure users higher than those from mixed users.
By introducing $A'$ and applying the augmentation methods, AFRR is decreased for mixed users and sometimes for pure users.
Our method (rightmost) improves the NDCG for the protected group by mainly raising the rank of spams from mixed users and sometimes the spams from pure users.

\subsubsection{Evaluation of the Joint method on improving the quality of $A'$.}
To answer \textbf{RQ2} and \textbf{RQ3}, we study the relationship between the group fairness and the quality of inferred $A'$.
Recall that in Table~\ref{tab: big table 20-percentile}, \ref{tab: big table 15-percentile}, and~\ref{tab: big table 10-percentile}, the \textbf{Joint} method has smaller $\Delta_{\text{NDCG}}$ than \textbf{Pre-trained} in most cases.
To understand the advantage of \textbf{Joint}, Fig.~\ref{fig: g_theta auc} shows the AUC gap of $A'$ given by $g_{\bt}$ ($x$-axis) in these two settings with corresponding $\Delta_{\text{NDCG}}$ difference ($y$-axis).
Most models in the area I indicate that \textbf{Joint} simultaneously promotes the accuracy of $g_{\bt}$ and the fairness of $f_{\bw}$.
Since \textbf{Joint} updates $\bt$ using the gradient from $f_{\boldsymbol{W}}$ (see Eq.~(\ref{eq: update theta})),
our mixup strategies can mitigate the overfitting for $g_{\boldsymbol{\theta}}$ with more gradients from the synthetic data.
\begin{figure*}
   \centering
   \vspace{-0.1cm}
    \subfigtopskip=2pt
    \subfigbottomskip = 2pt
    \subfigcapskip = -5pt
    \subfigure[20-th Percentile of User Node Degree]{
    \includegraphics[width=0.8\textwidth]{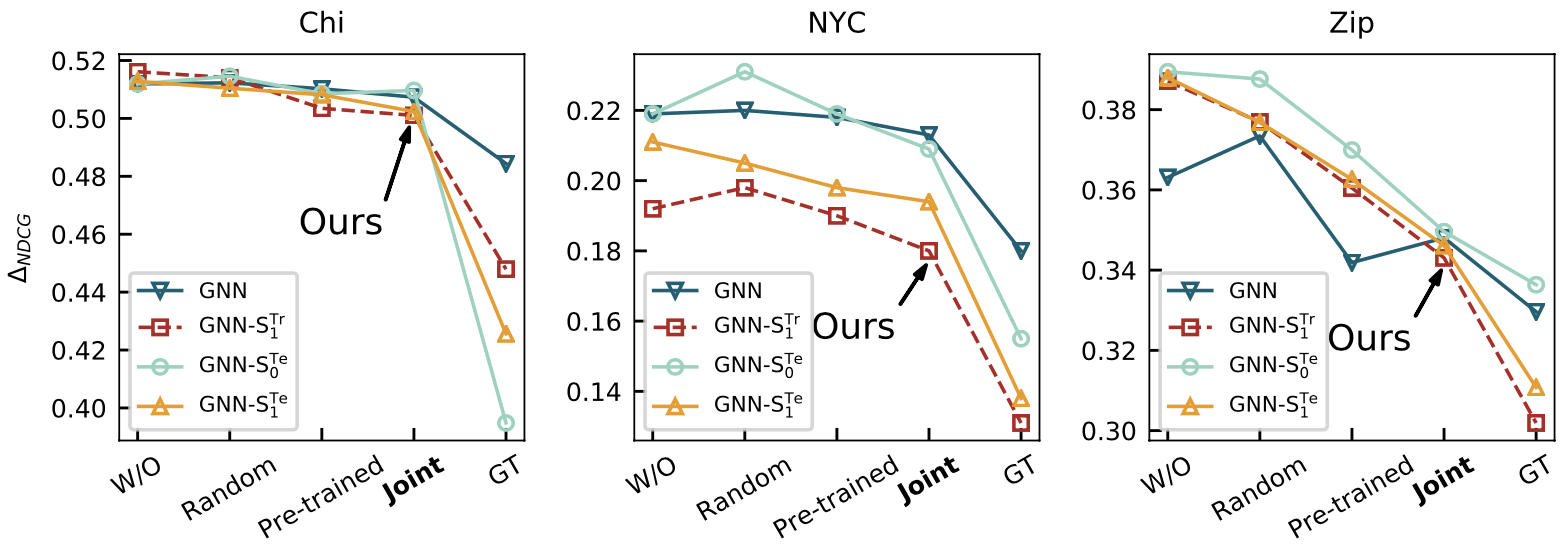}}
    \subfigure[15-th Percentile of User Node Degree]{
    \includegraphics[width=0.8\textwidth]{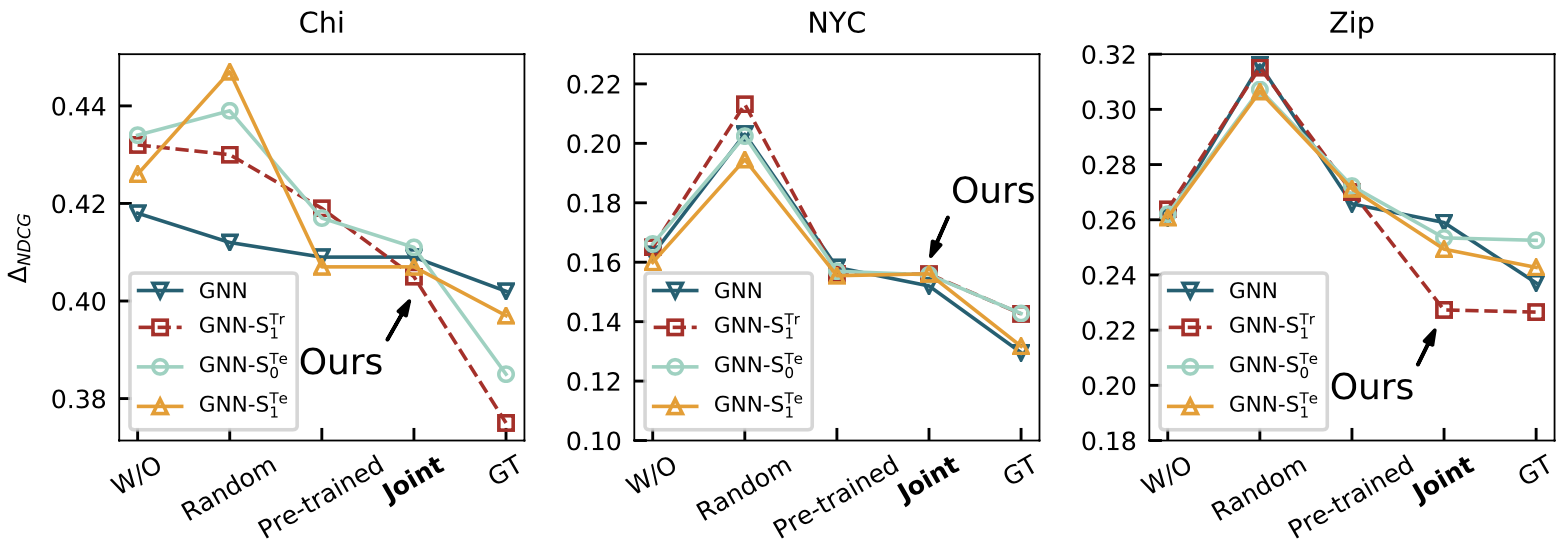}}
    \subfigure[10-th Percentile of User Node Degree]{
    \includegraphics[width=0.8\textwidth]{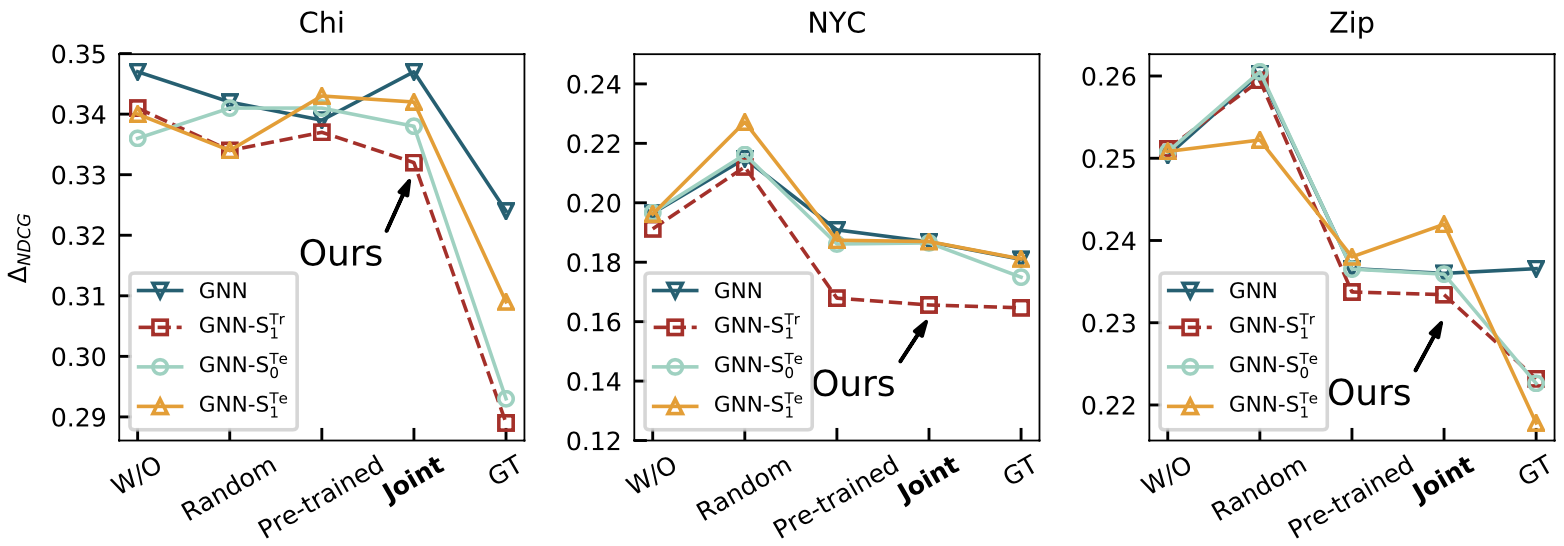}}
	\caption{Test $\Delta_{\text{NDCG}}$ for four detectors: GNN, GNN-$\text{S}_{1}^{\text{Tr}}$ (in dashed line; our method), GNN-$\text{S}_{0}^{\text{Te}}$, and GNN-$\text{S}_{1}^{\text{Te}}$ given $A'$ obtaining from five methods.
	The amount of noise in $A'$ reduces for methods from left to right: \textbf{w/o}: without $A'$; \textbf{Random}: randomly let $A'=1/0$; \textbf{Pre-trained}: output of pre-trained $g_{\boldsymbol{\theta}}$; \textbf{Joint}: output of jointly trained $g_{\boldsymbol{\theta}}$; \textbf{GT}: ground truth of $A'$).
	} 
    \label{fig: impact of A' accuracy.}
\end{figure*}
\subsubsection{Impact of the accurate $A'$ on group fairness.}
Since the quality of $A'$ correlates with the group fairness, we further increase (i.e., \textbf{Random} method) or decrease (i.e., \textbf{GT} method) the noise in $A'$ to see if this correlation holds to answer \textbf{RQ4}.
Fig.~\ref{fig: impact of A' accuracy.} gives $\Delta_{\text{NDCG}}$ for detectors with different ways to assign value to $A'$ which contain less and less noise moving from left to right in $x$-axis.
$\Delta_{\text{NDCG}}$ reduces as a detector receives a more accurate inference of the values of $A'$.

\subsection{Sensitivity studies for the replication times $k$ and if pruning non-spam edges.}
Fig.~\ref{fig: sensitive of duplication times k and pruning edges.} shows the test AUCs of $g_{\boldsymbol{\theta}}$ with replications $k=\{50, 100\}$ and if pruning non-spams (see Section~\ref{sec:minor_user_augmentation}).
Pruning has better AUCs than No pruning except when $k=100$ on Zip. 
Additionally, AUCs for Chi and Zip decrease as increasing the number of $k$.
Since Chi and Zip have few mixed users, forcing the synthesized data to mimic the original node distribution is complex and easy to cause $g_{\bt}$ overfitting.
Based on the validation set, we get the optimal $k=100$ for NYC and $k=50$ for Chi and Zip.
\section{Related Work}
\noindent\textbf{Fairness on graphs.}
People studied fairness on graphs from many perspectives.
In general, researchers attempt to obtain fair node embeddings, representations, or classification results towards some discrete and well-defined sensitive attributes~\cite{ijcai2019p456, buyl2020debayes}.
People introduce the adversarial framework to eliminate the unfairness bias concerning the sensitive attributes~\cite{dai2021say, bose2019compositional, agarwal2021towards}.
They added fairness regularizers to constrain the model insensitive for those attributes.
FairGNN~\cite{dai2021say} proposes an adversarial debiasing method with another GNN to estimate the missing sensitive information of the nodes.
Other works modify the graph topology structures or edge weights to obtain fair node embeddings or predictions.
FairAdj~\cite{li2021dyadic} adjusts the edge connections and learns a fair adjacency matrix by adding contains for graph structure.
FairEdit~\cite{loveland2022fairedit} proposes model-agnostic algorithms which perform edge addition and deletion by leveraging the gradient of fairness loss.
FairDrop~\cite{spinelli2021fairdrop} excludes the biased edges to counteract homophily.
Nevertheless, all these methods rely on known or well-defined sensitive attributes.
Additionally, some works study the fairness problem along with multiple sensitive attributes.
There will be some fairness violations once we force the model to be fair with respect to multiple sensitive attributes at the same time.
In~\cite{ustun2019fairness, kearns2018preventing}, researchers designed a classifier with multiple fairness constraints generated from the combinations of sensitive attributes.

\noindent\textbf{Augmentation on the graph.} 
Graph augmentation recently receives increasing attention~\cite{zhao2022graph}.
In~\cite{wang2021mixup, feng2020graph}, people studied the graph augmentation on the node level that synthetics data by mixup nodes or removing nodes from the original graph.
In addition, some work operates graph augmentation on the edge level where they modify (adding or removing edges) in either a deterministic~\cite{zhao2021data} or a stochastic~\cite{rong2019dropedge} way.
Furthermore, some methods augment graph data on the node feature level by randomly masking the node features~\cite{you2020graph}.
\section{Conclusion}
This work studies the fairness problem on a graph-based spam detection task, where the unfairness happens between the ``protected'' and ``favored'' groups defined by the known sensitive attribute node degree. 
To better describe the heterogeneous behaviors of ``favored'' users, $A'$ is defined to further split the ``favored'' users into the ``mixed'' and the ``pure'' users.  
To obtain the value of $A'$ for users with unobserved behaviors, i.e., test users, a second GNN $g_{\bt}$ is proposed to infer the value of $A'$ with our data augmentation methods.
We utilize the inferred $A'$ as an additional feature feeding into the GNN spam detector $f_{\bw}$.
Our \textbf{Joint} is designed to jointly train $f_{\bw}$ and $g_{\bt}$, allowing to enhance the fairness of the detector and improve the quality of inferred $A'$ simultaneously.
The results on three Yelp datasets demonstrate that \textbf{Joint} with two augmentation methods effectively promotes group fairness by increasing the suspiciousness of spam from both ``pure'' and ``mixed'' users.
\newpage
\bibliographystyle{splncs04}
\bibliography{paper}
\end{document}